\documentclass[sn-mathphys, Numbered, pdflatex]{sn-jnl}

\usepackage{graphicx}%
\usepackage{multirow}%
\usepackage{amsmath,amssymb,amsfonts}%
\usepackage{amsthm}%
\usepackage{mathrsfs}%
\usepackage[title]{appendix}%
\usepackage{xcolor}%
\usepackage{textcomp}%
\usepackage{manyfoot}%
\usepackage{booktabs}%
\usepackage{algorithm}%
\usepackage{algorithmicx}%
\usepackage{algpseudocode}%
\usepackage{listings}%
\usepackage{graphicx}
\usepackage{tabularray}
\usepackage{multirow}
\usepackage{colortbl}
\usepackage{booktabs}
\usepackage{hhline}
\usepackage{caption}
\usepackage{subcaption}
\usepackage{pifont}
\usepackage[normalem]{ulem}

\begin{document}

\title[Understanding the limitations of self-supervised learning for tabular anomaly 
detection]{Understanding the limitations of self-supervised learning for tabular anomaly 
detection}


\author*[1,2]{\fnm{Kimberly T.} \sur{Mai}}\email{kimberly.mai@ucl.ac.uk}

\author[2,3]{\fnm{Toby} \sur{Davies}}\email{t.davies@leeds.ac.uk}

\author[1]{\fnm{Lewis D.} \sur{Griffin}}\email{l.griffin@ucl.ac.uk}

\affil*[1]{\orgdiv{Department of Computer Science}, \orgname{University College London}, \orgaddress{\street{Gower Street}, \city{London}, \postcode{WC1E 6BT}, \country{United Kingdom}}}

\affil[2]{\orgdiv{Department of Security and Crime Science}, \orgname{University College London}, \orgaddress{\street{Gower Street}, \city{London}, \postcode{WC1E 6BT}, \country{United Kingdom}}}

\affil[3]{\orgdiv{School of Law}, \orgname{University of Leeds}, \orgaddress{\street{Woodhouse}, \city{Leeds}, \postcode{LS2 9JT}, \country{United Kingdom}}}


\abstract{While self-supervised learning has improved anomaly detection in computer vision and natural language processing, it is unclear whether tabular data can benefit from it. This paper explores the limitations of self-supervision for tabular anomaly detection. We conduct several experiments spanning various pretext tasks on 26 benchmark datasets to understand why this is the case.  Our results confirm representations derived from self-supervision do not improve tabular anomaly detection performance compared to using the raw representations of the data. We show this is due to neural networks introducing irrelevant features, which reduces the effectiveness of anomaly detectors. However, we demonstrate that using a subspace of the neural network's representation can recover performance.}


\maketitle
\textbf{Statements and Declarations}: This work was supported by funding from EPSRC under grant EP/R513143/1.

\section{Introduction}\label{sec1}

Anomaly detection is the task of identifying unusual instances. Two issues hinder performance: how to obtain a ``good" representation of the normal data and a lack of knowledge about the nature of anomalies. The emergence of self-supervised learning techniques has primarily addressed these issues in complex domains such as computer vision and natural language processing \cite{hendrycks2019ssl, mai2022nlp}. However, these techniques have not yielded the same benefits for tabular data \cite{reiss2022ad}.

Self-supervised learning typically uses a pretext task to learn the intrinsic structure of the training data \cite{balestriero2023ssl}. Examples of pretext tasks include colourising greyscale images \cite{zhang2016color} or predicting the next word in a sentence \cite{peters2018word, radford2018improving}. Understanding the typical characteristics of a domain allows one to choose an effective pretext task. For instance, colourisation requires knowledge of object boundaries and semantics. These aspects are useful for image classification \cite{geirhos2019texture, hermann2020texture}. However, unlike images or text where spatial or sequential biases are natural starting points for self-supervision, the starting points for tabular data are unclear.

A recent study indicated that self-supervised learning does not help tabular anomaly detection \cite{reiss2022ad}. Reiss et al. compared two self-supervised methods with $k$-nearest neighbours ($k$-NN) on the original features. Even though the methods were designed for tabular data, they found that $k$-NN on the original features worked the best. 

We seek to understand \textit{why} this is the case. We extend the experiments to include a more comprehensive suite of pretext tasks. We also incorporate synthetic test cases and analyse the underlying learnt representations. Our results reinforce that self-supervision does not improve tabular anomaly detection performance and indicate deep neural networks introduce redundant features, which reduces the effectiveness of anomaly detectors. Conversely, we can recover performance using a subspace of the neural network's representation. We also show that self-supervised learning can outperform the original representation in the case of purely localised anomalies and those with different dependency structures.

In addition to the above investigations, we ran a series of experiments to benchmark anomaly detection performance in a setting where we do not have access to anomalies during training. We include our findings as a complement to the self-supervision results and to provide practical insight into scenarios where specific detectors work better than others. 

Our contributions are as follows:
\begin{enumerate}
    \item We reconfirm the ineffectiveness of self-supervision for tabular anomaly detection.
    \item We empirically investigate why self-supervision does not benefit tabular anomaly detection.
    \item We introduce a comprehensive one-class anomaly detection benchmark using several self-supervised methods.
    \item We provide practical insights and identify instances where particular anomaly detectors and pretext tasks may be beneficial.
\end{enumerate}

In Section \ref{sec2}, we cover the anomaly detection setup. We proceed to outline our experimental approach in Section \ref{sec3}. We evaluate our findings in Section \ref{sec4}. Finally, we summarise our work and conclude in Section \ref{sec5}.

\section{Background}\label{sec2}

\subsection{Anomaly detection}
Anomaly detection can be characterised as follows:

Let $\mathcal{X} \in \mathbb{R}^d$ represent the data space.  We assume the normal data is drawn from a distribution $\mathcal{P}$ on $\mathcal{X}$. Anomalies are data points $\mathbf{x} \in \mathcal{X}$ that lie in a low probability region in $\mathcal{P}$. Therefore, the set of anomalies can be defined as follows \cite{ruff2021ad}:
\begin{equation}
    \mathcal{A} = \{\mathbf{x} \in \mathcal{X} | p(\mathbf{x}) \leq \tau\},\quad \tau \geq 0
\end{equation}

Where $\tau$ is a threshold. 
Often, the original input space is not used as anomaly detection performance can be improved by using a different representation space. In the context of deep learning, a neural network parameterised by $\theta: \mathcal{X} \mapsto \mathcal{Y}$ (where $\mathcal{Y} \in \mathbb{R}^m$) is used to transform the input data. The anomalies are assumed to lie in a low-probability region in the new space. Namely, $\mathcal{P}$ on $\mathcal{X}$ transforms to $\mathcal{P}'$ on $\mathcal{Y}$ according to $\mathcal{P}'(\theta(\mathbf{x})) = |\mathbf{J}_\theta|$, where $\mathbf{J}$ is the Jacobian of $\theta$. If $\theta$ is an effective mapping, then $\theta(\mathcal{A})$ will still be a low probability of $\mathcal{P}'$ and $\theta(\mathcal{A})$ will have a simpler boundary in $\mathcal{Y}$ than $\mathcal{A}$ in $\mathcal{X}$. 

There are deep anomaly detectors (which aim to simultaneously transform the data to a new subspace and classify it) and shallow anomaly detectors (which do not transform the data but solely rely on an existing representation). This paper focuses on shallow anomaly detectors to isolate the differences in representations derived from different self-supervision tasks. Evaluating the transformative properties of deep anomaly detectors is out of scope. In addition, recent approaches suggest state-of-the-art anomaly detection performance is achievable by separating the representation learning and detection components \cite{tack2020csi, sehwag2021ssd, reiss2021panda, sun2022knn, reiss2022ad}. In this setup, we also assume only normal samples are present in the training set. This is referred to as a ``one-class'' setting in anomaly detection literature. The expressions ``one-class learning" and ``anomaly detection" are synonymous  \cite{scholkopf1999ocsvm, chandola2009ad, ruff2021ad}. We use the same terms for consistency with the literature. We describe the anomaly detectors used in our analyses below. For a more detailed overview of anomaly detection techniques, we refer the reader to Ruff et al. \cite{ruff2021ad}.

\textbf{$k$-NN} assumes normal data closely surround other similar samples in the feature space, while anomalies have relatively fewer nearby neighbours. Despite being a simple approach, $k$-NN remains competitive in big data instances \cite{reiss2022ad, gu2019knn, bergman2020knn, reiss2021panda, sun2022knn}. $k$-NN typically uses features extracted from pre-trained classification neural networks \cite{bergman2020knn, reiss2021panda, sun2022knn} for image-based anomaly detection. However, equivalent neural networks for tabular data do not exist.

\textbf{Local outlier factor} (LOF) is a density-based outlier detection method \cite{breunig2000lof}. It compares the local density of a data point against its $k$-nearest neighbours. If the point's density is significantly lower, it is deemed anomalous.

\textbf{Isolation forest} (iForest) is an ensemble-based algorithm \cite{liu2008if}. It uses a set of isolation trees. Each tree aims to isolate the training data into leaves. The tree construction algorithm randomly selects an attribute and a random split inside the attribute's range until each data point lies in a leaf. Each observation is assigned a score by calculating the length of the root node to the leaf and averaging across the trees. Points with shorter path lengths are considered more unusual, as the algorithm assumes anomalies are easier to isolate.

\textbf{One-class support vector machine} (OCSVM) assumes normal data lies in a high-density region \cite{scholkopf1999ocsvm}. Taking the origin as an anchor in the absence of anomalous data during training, it learns a maximum margin hyperplane that separates most training data from the origin. The algorithm considers a test datum's distance to the learnt hyperplane to classify anomalies. The method classifies a point as an anomaly if it lies on the side of the hyperplane closer to the origin.

\textbf{Residual norms} belong to the category of dictionary-based approaches. Dictionary-based approaches assume the building blocks of a feature space can reconstruct normal data but cannot construct anomalies. Methods using dictionaries use either linear or non-linear manifold learning techniques (e.g., principal components analysis or autoencoders) to determine the building blocks \cite{huang2006pca, kim2020rapp, wang2022vim}. We use the linear principal space approach from Wang et al. \cite{wang2022vim} for our experiments. This technique achieves state-of-the-art results for out-of-distribution detection on images, verified in independent benchmarks \cite{yang2022ood}. Although introduced for images, the method itself is modality-neutral. We follow previous anomaly detection methodologies that have adapted image-based methods to other modalities while retaining acceptable performance \cite{reiss2022ad, mai2022nlp}. 

For consistency, we use Wang et al.'s \cite{wang2022vim} original notation and code implementation\footnote{\url{https://github.com/haoqiwang/vim}}. Given $\mathbf{X}$ as the in-distribution data matrix of training samples, we find the principal subspace $\mathbf{W}$ from the matrix $\mathbf{X}^T\mathbf{X}$. This subspace spans the eigenvectors of the $D$ largest eigenvalues of $\mathbf{X}^T\mathbf{X}$. We assume anomalies have more variance on the components with smaller explained variance \cite{kamoi2020mahalanobis}. Therefore, we project $\mathbf{X}$ to the subspace spanned by the \textit{smallest} eigenvalues of $D$ (represented by $\mathbf{W}^{\perp}$)  to encapsulate the residual space and take its norm as the anomaly score:

\begin{equation}
    ||\mathbf{x}^{W^\perp}||
\end{equation}

\subsection{An overview of self-supervised learning}

Self-supervision approaches devise tasks based on the intrinsic properties of the training data. By exploiting these properties, neural networks hopefully learn about the regularities of the data. Examples across different modalities include:

\textbf{Classifying perturbations}: Each training datum is subject to a perturbation randomly selected from a fixed set, such as rotating the input data \cite{gidaris2018rotnet} or reordering patches in an image \cite{noroozi2016jigsaw}. A classification model then learns to predict which perturbation was applied. 

\textbf{Conditional prediction}. A neural network sees pieces of the input data and learns to complete the remaining parts. Examples include predicting the next word given a portion of a sentence \cite{peters2018word} or filling in masked areas of an image \cite{pathak2016fill, he2022mae}. 

\textbf{Clustering}. Under this category, models learn to group semantically similar instances and place them far away from observations representing other semantic categories. $k$-means clustering is a classic example that measures similarity in Euclidean space. 

More modern techniques learn a similarity metric using neural mappings. One popular loss function that enables this is InfoNCE \cite{oord2016cpc, chen2020simclr}. InfoNCE takes augmented views of the same data point as positives and learns to group them while pushing away other data points. Variants of this method sample from the positive's nearest neighbours to create more semantic variations \cite{dwibedi2021nnclr, biswas2023pnnclr}. Augmentations are usually in the form of transformations. In the case of images, these can involve adding noise, colour jittering, or horizontal flips. However, InfoNCE relies on large batch sizes to enable sufficiently challenging comparisons. Augmentation choices are also vital, as aggressive transformations could remove relevant semantic features.

VICReg \cite{bardes2022vicreg} attempts to overcome some of the issues of InfoNCE by enforcing specific statistical properties. It encourages augmented views to have a high variance to ensure the neural mapping learns diverse aspects of the data. It also regularises the covariance matrix of the representations. This regularisation ensures the neural mapping covers complementary information across the representation space.

Additional pretext tasks are covered in more detail in Balestriero et al. \cite{balestriero2023ssl}. 

\subsection{Self-supervised learning and anomaly detection for non-tabular data}

Anomaly detection for non-tabular data has benefited from self-supervision. Golan and El-Yaniv \cite{golan2018geo} show that compared to OCSVMs trained on pixel space, outputs from a convolutional neural network trained to predict image rotations were more reliable for anomaly detection. Mai et al. \cite{mai2022nlp} demonstrate similar findings on text. They show that good anomaly detection performance is achievable by fine-tuning a transformer with a self-supervised objective and using the loss as an anomaly score.

Other successful approaches do not use a self-supervised model in an end-to-end manner for anomaly detection. The works of Sehwag et al. \cite{sehwag2021ssd} and Tack et al. \cite{tack2020csi} both extract features from neural networks trained with an InfoNCE objective to perform anomaly detection on images. Sehwag et al. classify anomalies using the Mahalanobis distance on the extracted space, while Tack et al. use a product of cosine similarities and norms. 

\subsection{Self-supervised learning and anomaly detection for tabular data}

Literature covering self-supervision for anomaly detection in tabular data is more limited. GOAD \cite{bergman2020goad} extends the work of Golan and El-Yaniv \cite{golan2018geo} to a more generalised setting. They apply random affine transformations to the data and train a neural network to predict these transformations. At inference, they apply all possible transformations to the test data, obtain the prediction of each transformation from the network and aggregate the predictions to produce the anomaly score. The network should be able to predict the correct modification with higher confidence for the normal data versus the anomalies.

ICL \cite{shenkar2022icl} adapts the InfoNCE objective. It considers one sample at a time. Taking a sample $\mathbf{x}_i$ of dimensionality $d$, ICL splits $\mathbf{x}_i$ into two parts. The dimensionality of the two parts depends on a given window size, $k$ $(k < d)$. The first part $\mathbf{a}_i$ is a continuous section of size $k$, while the second $\mathbf{b}_i$ is its complement of size $d - k$. A Siamese neural network containing two heads with dimensionalities $k$ and $d-k$ aims to push the representations together. The negatives are other contiguous segments of $\mathbf{x}_i$ of size $k$. As the neural network should be capable of aligning the normal data and not anomalies, the loss is the anomaly score.

Although both methods claim to be state-of-the-art for tabular anomaly detection, Reiss et al. \cite{reiss2022ad} did not find this to be the case. They replicated the pipelines of GOAD and ICL. In addition, they used the trained neural networks of GOAD and ICL as feature extractors. After extracting the features, they ran $k$-NN on the new representations. They compared both setups to $k$-NN on the original data. Although GOAD and ICL are specifically designed to process tabular data, Reiss et al. found that $k$-NN on the original data was the best-performing approach. However, they did not run a hyperparameter search to optimise the choice of $k$ (leaving it as $k=5$). They also used the original architectures designed for GOAD and ICL, which differ from each other. This choice could be another confounding factor affecting results.

We summarise the works that cover self-supervision and anomaly detection in Table \ref{tab:lit_summary}.

\begin{table}[!htbp]
\centering
\caption{Summary of related self-supervised anomaly detection literature across modalities.
\label{tab:lit_summary}}
\begin{tabular}{>{\hspace{0pt}}m{0.1\linewidth}>{\hspace{0pt}}m{0.058\linewidth}>{\hspace{0pt}}m{0.194\linewidth}>{\hspace{0pt}}m{0.242\linewidth}>{\hspace{0pt}}m{0.25\linewidth}} 
\toprule
\textbf{Modality}                                 & \textbf{Year} & \textbf{Author}    & \textbf{Pretext task}                                                              & \textbf{Anomaly detector}                 \\ 
\hline
\multirow{3}{0.092\linewidth}{\hspace{0pt}Images} & 2018          & Golan and El-Yaniv \cite{golan2018geo} & Rotation prediction                                                                & Classification confidence                 \\
                                                  & 2020          & Tack et al. \cite{tack2020csi}        & Contrastive learning                                                               & Cosine similarity and norm  \\
                                                  & 2021          & Sehwag et al. \cite{sehwag2021ssd}      & Contrastive learning                                                               & Mahalanobis distance                      \\ 
\hline
Text                                              & 2022          & Mai et al. \cite{mai2022nlp}         & Masked language modelling\par{}Causal language modelling\par{}Contrastive learning & Loss                                      \\ 
\hline
Tabular                                           & 2022          & Shenkar and Wolf \cite{shenkar2022icl}   & Contrastive learning                                                               & Loss                                      \\ 
\hline
\multirow{2}{0.092\linewidth}{\hspace{0pt}Multiple}  & 2020          & Bergman and Hoshen \cite{bergman2020goad} & Transformation prediction                                                          & Classification confidence                 \\
                                                  & 2022          & Reiss et al. \cite{reiss2022ad}       & Contrastive learning\par{}Transformation prediction                                & $k$-NN                                      \\
\bottomrule
\end{tabular}
\end{table}

\section{Method}\label{sec3}

\subsection{Datasets}
We use 26 multi-dimensional point datasets from Outlier Detection Datasets (ODDS) \cite{rayana2016odds}. Each datum comprises one record, which contains multiple attributes. Table \ref{tab1} summarises the properties of the datasets. We treat each dataset as distinct and train and test separate anomaly detection models for each dataset.

We follow the data split protocols described in previous tabular anomaly detection literature \cite{bergman2020goad, shenkar2022icl}. We randomly select 50\% of the normal data for training, with the remainder used for testing. The test split includes all anomalies. The training split did not use any anomalies as we adopt a one-class setup. We partition the training set further by leaving 20\% for validation. 

\begin{table}[h]
\caption{Summary of ODDS datasets.}\label{tab1}%
\begin{tabular}{llll} 
\toprule
\textbf{Dataset} & \textbf{Total size} & \textbf{Number of anomalies (\%)} & \textbf{Dimensionality}  \\ 
\midrule
Annthyroid       & 7,200               & 534 (7.4\%)                       & 6                        \\
Arrhythmia       & 452                 & 66 (14.6\%)                       & 274                      \\
BreastW          & 683                 & 239 (35.0\%)                      & 9                        \\
Cardio           & 1,831               & 176 (9.6\%)                       & 9                        \\
Glass            & 214                 & 9 (4.2\%)                         & 9                        \\
Heart            & 224                 & 10 (4.4\%)                        & 44                       \\
HTTP             & 567,469             & 2,211 (0.4\%)                     & 3                        \\
Ionosphere       & 351                 & 126 (35.8\%)                      & 33                       \\
Letter           & 1,600               & 100 (6.3\%)                       & 32                       \\
Lympho           & 148                 & 6 (4.1\%)                         & 18                       \\
Mammography      & 11,183              & 260 (2.3\%)                       & 6                        \\
MNIST            & 7,603               & 700 (9.2\%)                       & 100                      \\
Musk             & 3,062               & 97 (3.2\%)                        & 166                      \\
Optdigits        & 5,216               & 150 (2.9\%)                       & 64                       \\
Pendigits        & 6,870               & 156 (2.3\%)                       & 16                       \\
Pima             & 768                 & 268 (34.9\%)                      & 8                        \\
Satellite        & 6,435               & 2,036 (31.6\%)                    & 36                       \\
Satimage-2       & 5,803               & 71 (1.2\%)                        & 36                       \\
Seismic          & 2,584               & 170 (6.5\%)                       & 11                       \\
Shuttle          & 49,097              & 3,511 (6.6\%)                     & 9                        \\
SMTP             & 95,156              & 30 (0.03\%)                       & 3                        \\
Speech           & 3,686               & 61 (1.7\%)                        & 400                      \\
Thyroid          & 3,772               & 93 (2.4\%)                        & 6                        \\
Vertebral        & 240                 & 30 (12.5\%)                       & 6                        \\
Vowels           & 1,456               & 50 (3.4\%)                        & 12                       \\
WBC              & 278                 & 21 (5.6\%)                        & 30                       \\
Wine             & 129                 & 10 (7.7\%)                        & 13                       \\
\bottomrule
\end{tabular}
\end{table}

\subsection{Baseline approach}

We run $k$-NN, iForest, LOF, OCSVM, and residual norms on the original training data. As we aim to expand on the work of Reiss et al. \cite{reiss2022ad}, we only implement one-class detectors for comparability. Even though Reiss et al. \cite{reiss2022ad} only use $k$-NN in their experiments, we use multiple detectors to establish whether $k$-NN is the best detector or if there are other more appropriate detectors depending on the type of anomalies present.  We analyse our findings in Section \ref{sec4:benchmark}. Another anomaly detection study, ADBench \cite{han2022adbench}, follows a similar protocol. However, their setup assumes anomalies are present in the training data. Through our experiments, we establish whether a purely one-class setup affects overall detector ranking. We use scikit-learn \cite{pedregosa2011scikit} to implement all detectors except for $k$-NN, which uses the faiss library \cite{johnson2021faiss}.

We also investigate the detectors' sensitivity to different configurations by varying the hyperparameters. For $k$-NN and LOF, we report results for $k=\{1,2,5,10,20,50\}$. For the residual norms, we look at how results change with a proportion of features, with percentages ranging from 10\% to 90\% in 10\% increments $[10\%, 20\%, ..., 90\%]$. We record our findings in Section \ref{sec4:benchmark}. For the self-supervised tasks, we report the results based on the best hyperparameter configuration derived from these ablations. We retain the default scikit-learn parameters for iForest and OCSVM, which uses a radial basis function kernel. 

The detectors run directly on the data and on a standardised version. We standardise each dimension independently by removing the mean and scaling to unit variance. We also experimented with fully whitening the data but found attribute-wise standardisation rendered similar results.

\subsection{Self-supervision}

\subsubsection{Pretext tasks}
Although tabular data lacks overt intrinsic properties like those in images or text, we choose self-supervised tasks that we hypothesise can take advantage of its structure.

Firstly, we adapt ICL \cite{shenkar2022icl} and GOAD \cite{bergman2020goad} to use them as pretext tasks. We do not directly implement  ICL and GOAD as they score anomalies in an end-to-end manner. In contrast, our experiments focus on how representations from different pretext tasks affect shallow detection performance. Therefore, we refer to the ICL-inspired task as ``\textbf{EICL}'' (embedding-ICL) for the remainder of the paper. As GOAD uses random affine transformations, we can consider this a combination of predicting rotation and stretches. This configuration conflates two different tasks and could be trivial to solve. Therefore, we attempt to align it closer to the RotNet \cite{ hendrycks2019ssl, gidaris2018rotnet} experiments for image-based anomaly detection by training a model to classify orthonormal rotations. This pretext task should profit from the rotationally invariant property of tabular data \cite{grinsztajn2022tree}. Hence we refer to the GOAD-inspired task as ``\textbf{Rotation}''.

The additional objectives used in the experiments are as follows:

\textbf{Predefined shuffling prediction (Shuffle)}: We pick a permutation of the dimensions of the data from a fixed set of permutations and shuffle the order of the attributes based on the selection. The model learns to predict that permutation.

\textbf{Predefined mask prediction (Mask classification)}: Given a mask rate $r\quad(r < d)$, we initialise predefined classes that indicate which attributes to mask. We perform masking by randomly selecting another sample $\mathbf{x_j}$ from the training set and replacing the chosen attributes in $\mathbf{x_i}$ with those from $\mathbf{x_j}$. We follow the protocol outlined in Yoon et al. \cite{yoon2020vime} This approach generated better representations compared to alternative masking strategies like imputation, and constructing a mask classification pretext task outperformed alternative supervised and semi-supervised methods on tabular classification tasks. The model learns to classify which predefined class was applied.

\textbf{Masked columns prediction (Mask columns)}: The model picks which attributes were masked given a mask rate $r$. For example, if only the first attribute was masked, a correct classification should identify the first attribute and should not pick the other attributes. This is different from the mask classification task, where the predefined mask class is given a label from a fixed set of combinations rather than from the particular attribute that has been masked (for example, if there are only two classes, the labels for mask classification are 0 or 1). 

\textbf{Denoising autoencoding (Autoencoder)}: Given a mask rate $r$, we perturb $\mathbf{x_i}$ by randomly selecting another sample $\mathbf{x_j}$ and replacing a subset of $\mathbf{x_i}$'s attributes with those of $\mathbf{x_j}$. The perturbed $\mathbf{x_i}$ is the input. Given this input, the model learns to reconstruct the unperturbed $\mathbf{x_i}$.

\textbf{Contrastive learning}: We create positive views of $\mathbf{x_i}$ by rotating the data using an orthonormal matrix \textbf{(Contrastive rotation)}, permuting the attributes per the shuffle task \textbf{(Contrastive shuffle)}, or masking the attributes per the mask classification task \textbf{(Contrastive mask)}. We treat other data points in a minibatch as negatives. We only apply one augmentation at a time to isolate their effects.

\subsubsection{Network architectures and loss functions}
We use the same neural network architectures to control for any potential effects on performance. Per the findings of Gorishniy et al. \cite{gorishniy2021tab}, we use ResNets \cite{he2016resnet} and FT-Transformers. Gorishniy et al. examined the performance of several deep learning architectures on tabular classification and regression, including multilayer perceptrons, recurrent neural networks, ResNets and transformers. Their results indicated that ResNets and FT-Transformers were the best overall. Based on these findings, we restrict our architectures to the most promising variants. FT-Transformer is a transformer specially adapted for tabular inputs where each transformer layer operates on the feature level of one datum. 

We train both architectures on all objectives except for EICL, where we only use ResNets. As EICL requires specific partitioning of the features, the FT-Transformer architecture would need to be modified. This modification is out of the scope of our experiments. We retain the same architecture (e.g., the number of blocks) for each pretext task and only vary the dimensionality of the output layer. The dimensionality corresponds to the number of preset classes for the rotation, shuffle, and mask classification tasks. The output dimensionality of the autoencoder task mirrors the input dimensionality. For the contrastive objectives (including EICL), we set the output as one of $\{128, 256, 512\}$ depending on validation performance.

As previous literature has claimed specialised loss functions can improve out-of-distribution detection on other modalities \cite{chen2022arpl,vaze2022ood}, we examine these to confirm whether they also improve tabular anomaly detection. 

For the rotation, shuffle, and mask classification tasks, we use cross-entropy, adversarial reciprocal points learning (ARPL) \cite{chen2022arpl}, and additive angular margin (AAM) \cite{deng2022arc}. ARPL is a specialised loss function for out-of-distribution detection. The probability of a datum belonging to a class is proportional to its distance to a reciprocal point. The point represents ``otherness'' in the learnt feature space. AAM is a loss function typically used for facial recognition. AAM specifically enforces interclass similarity and ensures interclass separation using a specified margin. This results in more spherical features for each class. We include AAM as some literature claims spherical per-class features make out-of-distribution detection easier \cite{ming2023cider}. Finally, we incorporate the cross-entropy loss as studies have shown models trained with this loss function can meet or outperform specialised losses like ARPL with careful hyperparameter selection \cite{vaze2022ood}. We experiment with mean squared error and mean absolute error for the autoencoders. We use the binary cross-entropy loss for masked column prediction, as multiple masked columns correspond to more than one label for each datum. For the contrastive objectives, we experiment with both InfoNCE and VICReg.

We summarise all the possible model configurations in Table \ref{tab2}.

\begin{table}[h!]
\centering
\caption{Summary of the model configurations.\label{tab2}}
\setlength{\extrarowheight}{0pt}
\addtolength{\extrarowheight}{\aboverulesep}
\addtolength{\extrarowheight}{\belowrulesep}
\setlength{\aboverulesep}{0pt}
\setlength{\belowrulesep}{0pt}
\arrayrulecolor{black}
\begin{tabular}{llll} 
\toprule
\textbf{Anomaly detectors}                                                                                                                                                                                                                       & \textbf{~Architectures}                                                         & \textbf{Self-supervised tasks}                                             & \textbf{Loss functions}                                                                                                                  \\ 
\hline
{\cellcolor[rgb]{0.929,0.929,0.929}}                                                                                                                                                                                                             & \multirow{9}{*}{\begin{tabular}[c]{@{}l@{}}ResNet\\FT-Transformer\end{tabular}} & {\cellcolor[rgb]{0.929,0.929,0.929}}Rotation               & \multirow{3}{*}{\begin{tabular}[c]{@{}l@{}}Cross-entropy\\ARPL\\AAM\end{tabular}}  \\
{\cellcolor[rgb]{0.929,0.929,0.929}}                                                                                                                                                                                                             &                                                                                 & {\cellcolor[rgb]{0.929,0.929,0.929}}Shuffle                                &                                                                                                                                          \\
{\cellcolor[rgb]{0.929,0.929,0.929}}                                                                                                                                                                                                             &                                                                                 & {\cellcolor[rgb]{0.929,0.929,0.929}}Mask classification                        &                                                                                                                                          \\ 
\hhline{>{\arrayrulecolor[rgb]{0.929,0.929,0.929}}-~->{\arrayrulecolor{black}}-}
{\cellcolor[rgb]{0.929,0.929,0.929}}                                                                                                                                                                                                             &                                                                                 & {\cellcolor[rgb]{0.929,0.929,0.929}}Mask columns                           & Binary cross-entropy                                                                                                                     \\ 
\hhline{>{\arrayrulecolor[rgb]{0.929,0.929,0.929}}-~->{\arrayrulecolor{black}}-}
{\cellcolor[rgb]{0.929,0.929,0.929}}                                                                                                                                                                                                             &                                                                                 & {\cellcolor[rgb]{0.929,0.929,0.929}}Autoencoder                            & \begin{tabular}[c]{@{}l@{}}MSE\\MAE\end{tabular}                                                                                         \\ 
\hhline{>{\arrayrulecolor[rgb]{0.929,0.929,0.929}}-~->{\arrayrulecolor{black}}-}
{\cellcolor[rgb]{0.929,0.929,0.929}}                                                                                                                                                                                                             &                                                                                 & {\cellcolor[rgb]{0.929,0.929,0.929}}EICL                                    & \multirow{4}{*}{\begin{tabular}[c]{@{}l@{}}InfoNCE\\VICReg\end{tabular}}                                                                 \\
{\cellcolor[rgb]{0.929,0.929,0.929}}                                                                                                                                                                                                             &                                                                                 & {\cellcolor[rgb]{0.929,0.929,0.929}}Contrastive - rotation    &                                                                                                                                          \\
{\cellcolor[rgb]{0.929,0.929,0.929}}                                                                                                                                                                                                             &                                                                                 & {\cellcolor[rgb]{0.929,0.929,0.929}}Contrastive - shuffle &                                                                                                                                          \\
\multirow{-9}{*}{{\cellcolor[rgb]{0.929,0.929,0.929}}\begin{tabular}[c]{@{}>{\cellcolor[rgb]{0.929,0.929,0.929}}l@{}}$k$-nearest neighbours\\Isolation forest\\Local outlier factor\\One-class support vector machine\\Residual norms\end{tabular}} &                                                                                 & {\cellcolor[rgb]{0.929,0.929,0.929}}Contrastive - mask        &                                                                                                                                          \\
\bottomrule
\end{tabular}
\end{table}

\subsubsection{Model selection}
Due to the number of potential hyperparameter combinations, we perform random searches to determine the most appropriate models for anomaly detection. We pick hyperparameters randomly and train on the training split for each self-supervised task and dataset. As we cannot evaluate using anomalies, we select models that achieve the lowest loss on the normal validation data. As we want to analyse the effect of different loss functions and architecture, the hyperparameter sweep stage results in a maximum of twelve configurations for each dataset and task. For example, the models trained on the rotation task would include ResNets and FT-Transformers, each architecture also includes the cross-entropy, ARPL, and AAM losses. There are also different configurations for standardised and non-standardised input data.

\subsubsection{Feature extraction}
\begin{figure}[h]
\centering
\includegraphics[width=\linewidth]{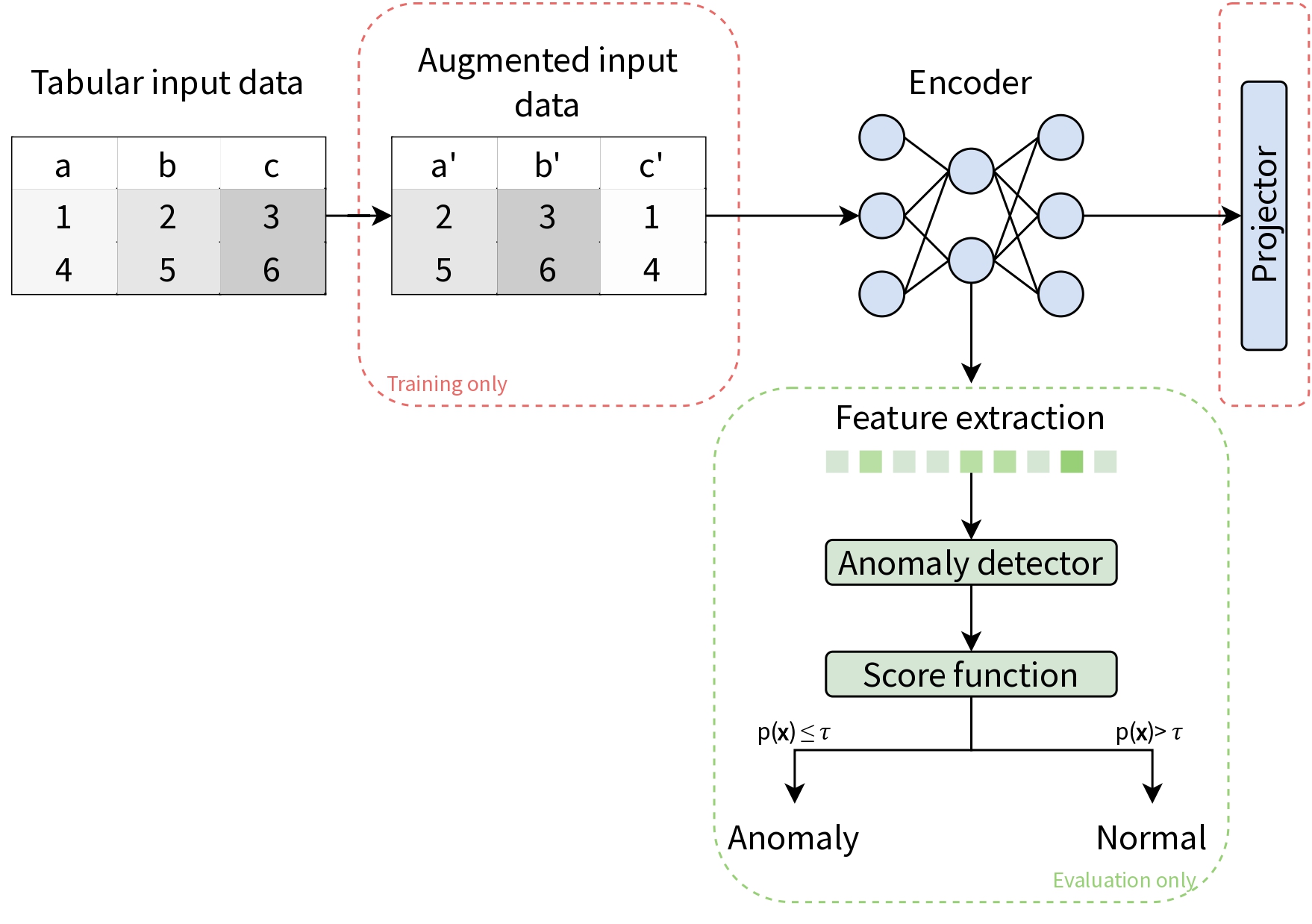}
\caption{Self-supervised anomaly detection workflow. The data are only augmented and fed through the projector during training.}
\label{workflow}
\end{figure}

After training, we obtain the learnt features by passing input data through the self-supervised models. We extract the features from the penultimate layer. As we fix the architecture for the different tasks, we obtain 128-dimensional embeddings for ResNets and 192-dimensional embeddings for FT-Transformer. We train the anomaly detectors using the new training features and test them using the transformed test features. We do not apply any augmentations during inference to ensure a fair comparison between the self-supervised tasks. Figure \ref{workflow} shows the workflow.

\subsection{Evaluation}

We evaluate all anomaly detectors using the area under the receiving operator curve (AUROC) score. We can consider AUROC as the probability that a randomly selected anomaly will be ranked as more abnormal than a normal sample. Scores fall between 0\% and 100\%. A score of 50\% indicates the detector cannot distinguish between anomalies and normal data points, while a score of 100\% signals perfect anomaly discrimination. We choose AUROC as it does not require a threshold to control for false positives, for example. 

\subsection{Additional ablations}
In addition to evaluations with the ODDS dataset, we run more experiments to understand detector performance and scenarios where specific self-supervised objectives may perform better than others.

\subsubsection{Synthesised anomalies}
Although ODDS contains several datasets, the datasets may mix different types of anomalies. These mixes can make it difficult to diagnose why one representation performs better than another. Therefore, we evaluate how the pretext tasks and their learnt representations fare with synthesised anomalies. We keep the normal data in the train and test splits and only generate anomalies by perturbing the properties of the normal training data. We use the four synthetic anomaly categories as defined in ADBench \cite{han2022adbench, steinbuss2021ad}. We use ADBench's code to create all types.
\begin{itemize}
    \item \textbf{Local} anomalies deviate from their local cluster. We use Gaussian mixture models (GMM) to learn the underlying normal distribution. The covariance matrix undergoes scaling by a factor $\alpha$ to generate the anomalies. We use $\alpha = 2$ in our experiments.
    \item \textbf{Cluster} anomalies use GMMs to learn the normal distribution. A factor $\beta$ scales the mean feature vector to create the cluster anomalies. We use $\beta = 2$ in our experiments.
    \item \textbf{Global} anomalies originate from a uniform distribution $U [ \delta \cdot \textrm{min}(\mathbf{X}_i^k), \delta \cdot \textrm{max}(\mathbf{X}_i^k)]$. $\delta$ is a scaling factor, and the minimum and maximum values of an attribute $\mathbf{X}_i^k$ define the boundaries. We use $\delta = 0.01$.
    \item \textbf{Dependency} anomalies do not follow the regular dependency structure seen in normal data. We use vine copulas to learn the normal distribution and Gaussian kernel density estimators to generate anomalies.
\end{itemize}

\subsubsection{Corrupted input data}

Previous work hypothesises neural networks underperform on tabular classification and regression because of their rotational invariance and lack of robustness to uninformative features \cite{grinsztajn2022tree}. We investigate if this occurs for anomaly detection. Simultaneously, we explore the shallow anomaly detectors' sensitivity to corrupted attributes. Understanding these results can give a practical insight into what self-supervision objectives and anomaly detectors work best when the data is noisy or incomplete. For our ablations, we follow Grinsztajn et al. \cite{grinsztajn2022tree} and apply the following corruptions to the raw data:

\begin{enumerate}
    \item \textbf{Adding uninformative features}: We add extra attributes to $\mathbf{X}$. We select a subset of attributes to imitate. We then generate features by sampling from a multivariate Gaussian based on the mean and interquartile range of the subset's values. We experiment with different proportions of additional features and limit the maximum number of extra attributes to be no greater than the existing number of features in the dataset.
    \item \textbf{Missing values}:  We randomly remove a proportion of the entries and replace the missing values using the mean of the attribute the value belongs to. We apply this transformation to both the train and test sets.
    \item \textbf{Removing important features}: We train a random forest classifier to classify between normal samples and anomalies. We then drop a proportion of attributes based on the feature importance values output by the random forest, starting from the least important. This corruption violates the one-class assumption within our anomaly detection setup. However, we use this to analyse the robustness of the detectors and self-supervised models.
    \item \textbf{Selecting a subset of features}: Similar to (3), we train a random forest classifier. We choose a proportion of attributes based on the feature importance values output from the random forest, starting from the most important. 
\end{enumerate}

After corrupting the data, we follow the same process of training the self-supervised models and feature extraction for the neural network experiments.
\clearpage
\section{Results}\label{sec4}

We organise our results as follows: Subsection \ref{sec4_overall} reconfirms the ineffectiveness of self-supervision for tabular anomaly detection and summarises the main results at a high level. We investigate this phenomenon through a series of case studies and ablations. Subsections \ref{sec4_http} and \ref{sec4_toy} drill down on performance using a subset of ODDS (\textit{HTTP}) and simplified toy scenarios. Our working hypothesis is that self-supervision introduces irrelevant directions. We empirically verify our hypothesis by investigating the residual space of the embeddings in Subsection \ref{sec4:oddsanalysis}. We attempt to compare the properties of the self-supervised pretext tasks by replacing ODDS anomalies with synthetic variants in Subsection \ref{sec4_synthetic}. Finally, we investigate the effect of architecture and detector choices in Subsection \ref{sec4_architecture}.

\subsection{Self-supervision results}
\label{sec4_overall}

\begin{figure}[!htbp]
    \centering
    \includegraphics[width=0.8\linewidth]{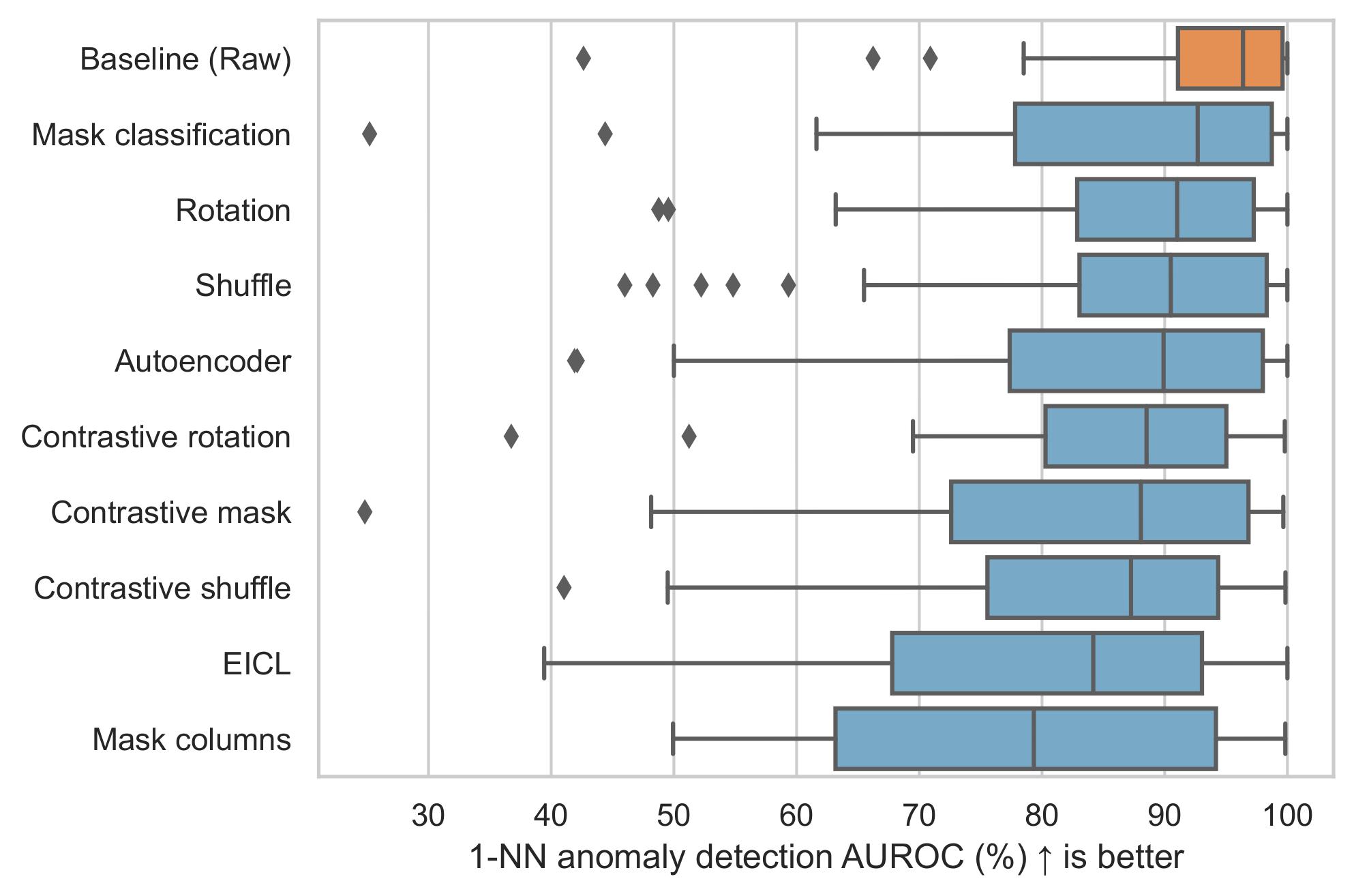}
    \caption{Box plot comparing nearest neighbour AUROCs for each of the embeddings, ordered by median performance. For each self-supervised task, we filter the results by architecture and loss function to include the embedding with the best-performing results.}
    \label{fig:ssl_boxplot}
\end{figure}

\begin{figure}[!htbp]
    \centering
    \includegraphics[width=0.8\linewidth]{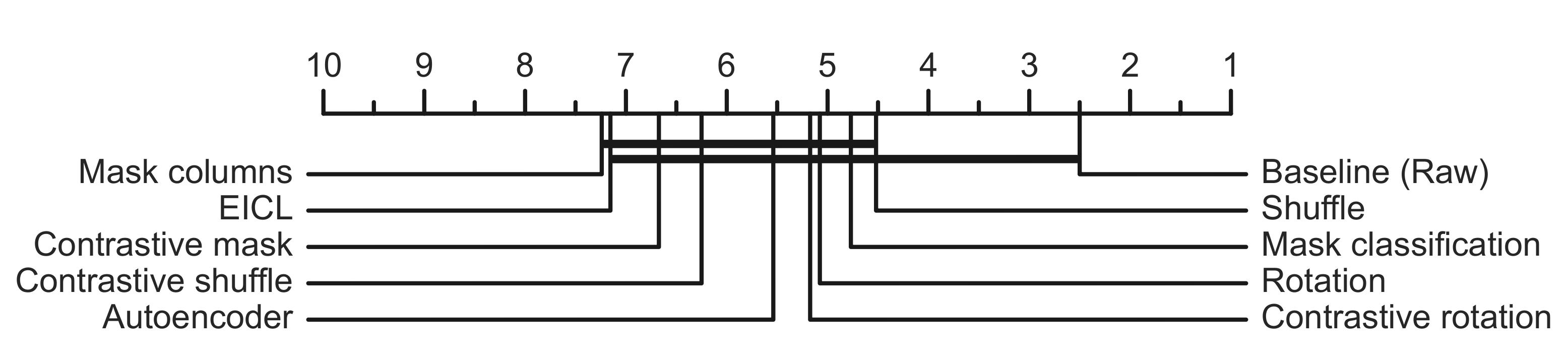}
    \caption{Critical difference diagram comparing the embeddings in a pairwise manner. The horizontal scale denotes the average rank of each embedding. The dark lines between different detectors indicate a statistical difference ($p < 0.05$) in results when running pairwise comparison tests. The baseline scores greatly outrank the pretext tasks. In contrast, the scores among the pretext tasks are more closely aligned.}
    \label{fig:ssl_critical}
\end{figure}

\textbf{No self-supervision task outperforms the baseline}. Figure \ref{fig:ssl_boxplot} summarises the nearest neighbour performance derived from the embeddings of each self-supervised approach. We aggregate performance by representation rather than dataset to concentrate on the influence each representation has on performance. No self-supervision task exceeds $k$-NN on the raw tabular data. When comparing results at a pairwise level, Figure \ref{fig:ssl_critical} shows that the baseline scores greatly outrank the self-supervised objectives. Similarly, performance using the self-supervised embeddings drops in the presence of corrupted data (Appendix, Figure \ref{fig:corrupt_ssl}). These results extend the findings in \cite{grinsztajn2022tree} that neural networks are also more sensitive to corrupted attributes in the anomaly detection task. When excluding the baseline, the classification-based tasks (shuffle, mask classification, and rotation) outperform their contrastive and reconstructive counterparts. 

\begin{figure}[h!]
    \centering
    \includegraphics[width=0.8\linewidth]{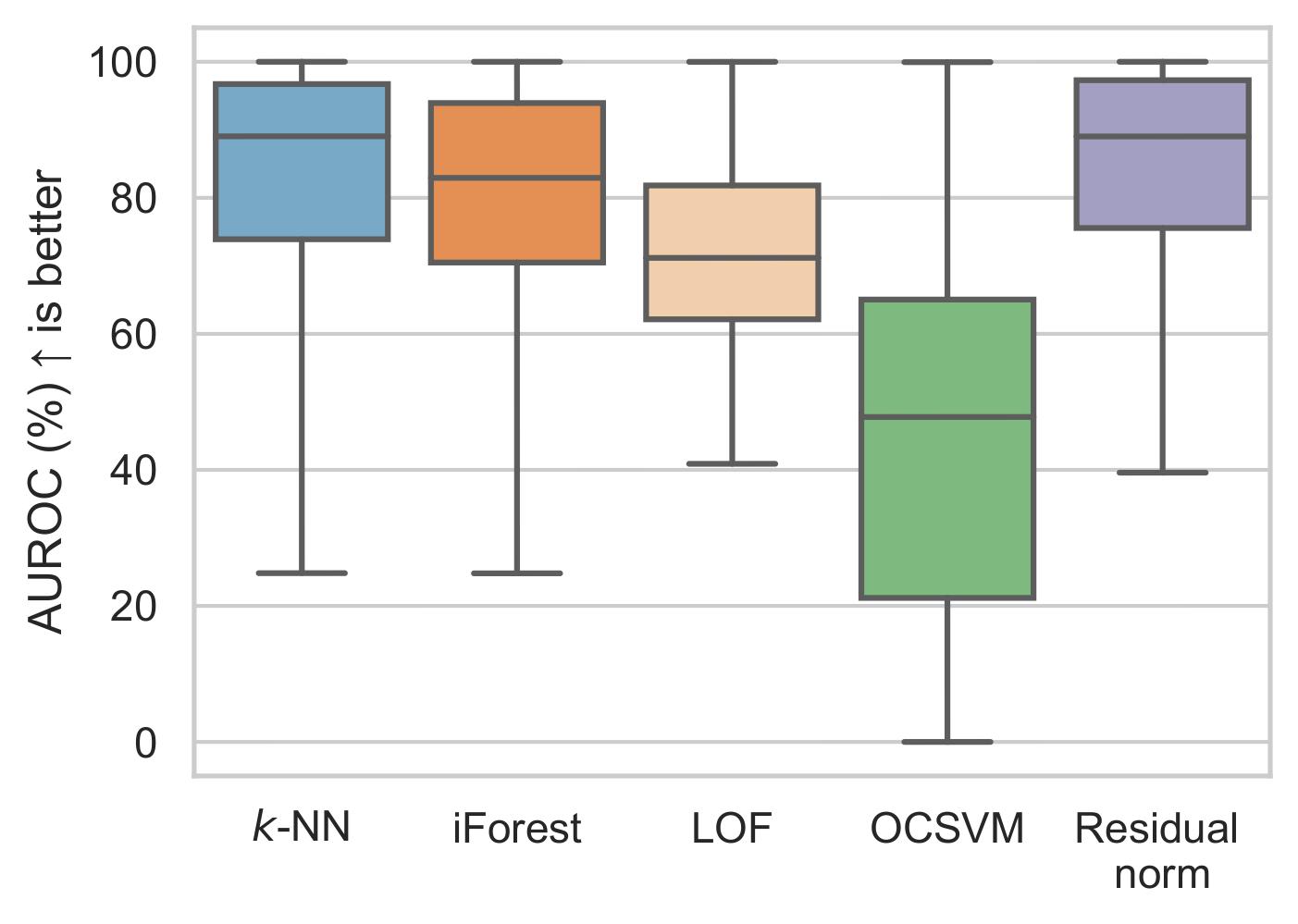}
    \caption{Box plot comparing detector performance on the self-supervised embeddings.}
    \label{fig:ssl_detector}
\end{figure}

We observe similar results when we use different shallow detectors to perform anomaly detection (Figure \ref{fig:ssl_detector}), with one exception. Using residual norms on the embedding space is a better choice than $k$-NN. However, they still lag behind $k$-NN scores on the original embeddings. We also observe that OCSVM performs consistently worse across all tasks.

\subsection{A case study on HTTP}
\label{sec4_http}

To understand why self-supervision does not help, we will explore one ODDS dataset in detail. We proceed to test our reasoning on toy datasets and then analyse the remaining ODDS datasets.

We use \textit{HTTP} for our analyses. \textit{HTTP} is a modified subset of the KDD Cup 1999 competition data \cite{misc_kdd_cup_1999_data_130}. The competition task involved building a detector to distinguish between intrusive (attack) and typical network connections. The dataset initially contained 41 attributes from different sources, including HTTP, SMTP, and FTP. The ODDS version only uses the ``service'' attribute from the \textit{HTTP} information as it is considered one of the most basic features. The resulting subset is three-dimensional and comprises over 500,000 observations. Out of these samples, 2,211 (0.4\%) are attacks.

It is easy to find attacks when running detectors directly on the raw ODDS variant of \textit{HTTP}. In our experiments, all shallow methods achieve AUROCs between 87.9\% and 100\% on non-standardised data, with the median score being 99.7\%. Further investigations show the attacks are separate from typical connections. A supervised logistic regression model trained to classify the two classes achieves 99.6\% AUROC, even with only 200 sample anomalies for training. 

\begin{figure}[h!]
    \centering
    \includegraphics[width=0.8\linewidth]{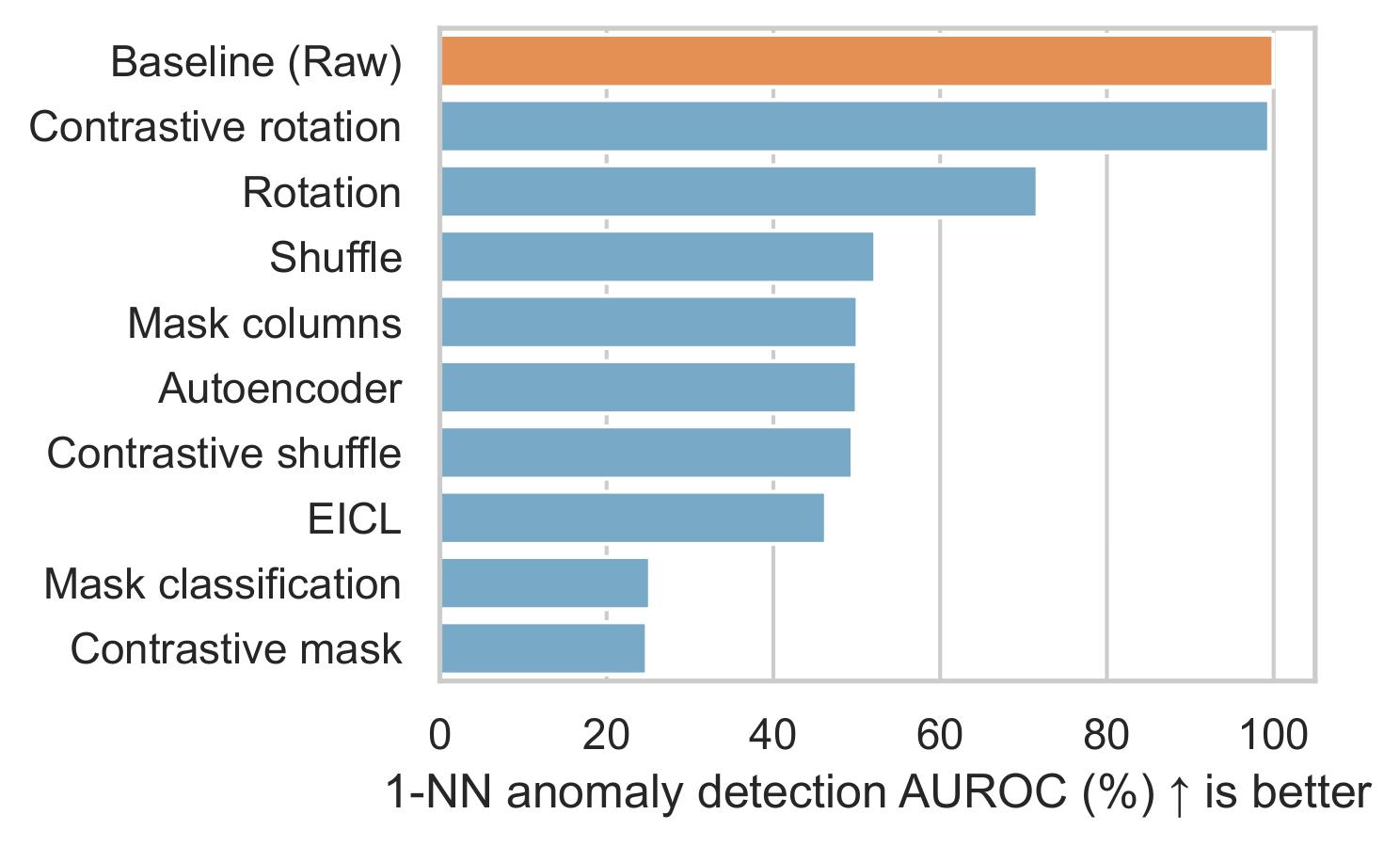}
    \caption{Bar chart comparing baseline and self-supervised embedding results on HTTP.}
    \label{fig:http}
\end{figure}
However, we observe peculiar results when using representations devised from the pretext tasks for \textit{HTTP}. $k$-NN performance drops drastically across the majority of tasks (Figure \ref{fig:http}), sometimes yielding scores worse than random. Conversely, the other detectors maintain their performance. For example, when extracting features from the rotation task\footnote{Using the best-performing rotation model, which is an FT-Transformer trained with ARPL loss.}, $k$-NN obtains 71.8\% AUROC, while iForest, OCSVM, and residual norms preserve AUROCs around 99\%. In addition, logistic regression continues to classify anomalies with 99\% AUROC in the supervised setting using the rotation task representations. As $k$-NN is susceptible to the curse of dimensionality, these initial results suggest the neural network representation introduces directions that obscure informative distances between the typical and intrusive samples.  Moreover, as iForest uses a splitting strategy for detection, its consistent results indicate \textit{some} direction signalling anomalousness exists.

\subsection{Toy data analysis}
\label{sec4_toy}

It can be challenging to draw conclusions based on existing datasets, as they are large and often contain uninterpretable features. Therefore, we pivot to toy examples to understand these behaviours. We devise nine two-dimensional toy datasets of varying difficulty (Appendix, Figure \ref{fig:toy_vis}). Like the experiments on the ODDS, we first evaluate performance directly on the two-dimensional representations. We then train ResNets on a two-class rotation prediction task, extract features from the penultimate embedding and re-run the detectors on the new space. We use this setting as rotations can be performed on two-dimensional data, and ResNets require less compute than the FT-Transformers.  We apply the same architecture as the ODDS experiments, making the extracted features 128-dimensional.

\begin{figure}[h!]
    \centering
    \begin{subfigure}[b]{0.3\linewidth}
         \centering
         \includegraphics[width=\textwidth]{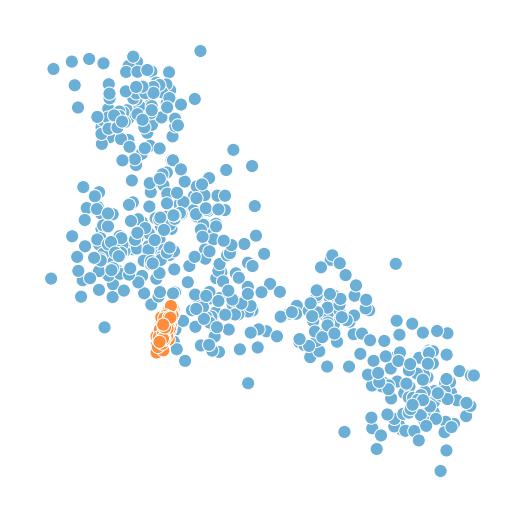}
         \caption{Original 2D data}
     \end{subfigure}
     \quad
     \begin{subfigure}[b]{0.3\linewidth}
        \centering
         \includegraphics[width=\textwidth]{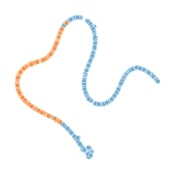}
         \caption{t-SNE visualisation of the self-supervised features}
     \end{subfigure}
    \caption{Visualisations of the multiple Gaussian toy dataset. Light blue are the normal data and orange are the anomalies. The features extracted from the neural network appear to be more narrow (b) and stretched compared to their original 2D representation (a).}
    \label{fig:toy_example}
\end{figure}
Regardless of whether the network can or cannot identify the rotation applied to the data, we observe behaviours consistent with ODDS in most toy instances. Compared to the original two-dimensional results, detection performance drops for almost all detectors after extracting representations from the ResNets. As two dimensions are sufficient to capture the characteristics of the datasets, projecting the data to a 128-dimensional space only results in a stretched and narrow representation without extra information. The t-SNE plots highlight this activity. We show an example of the multiple Gaussian dataset in Figure \ref{fig:toy_example}.

\begin{figure}[h!]
    \centering
    \includegraphics[width=\linewidth]{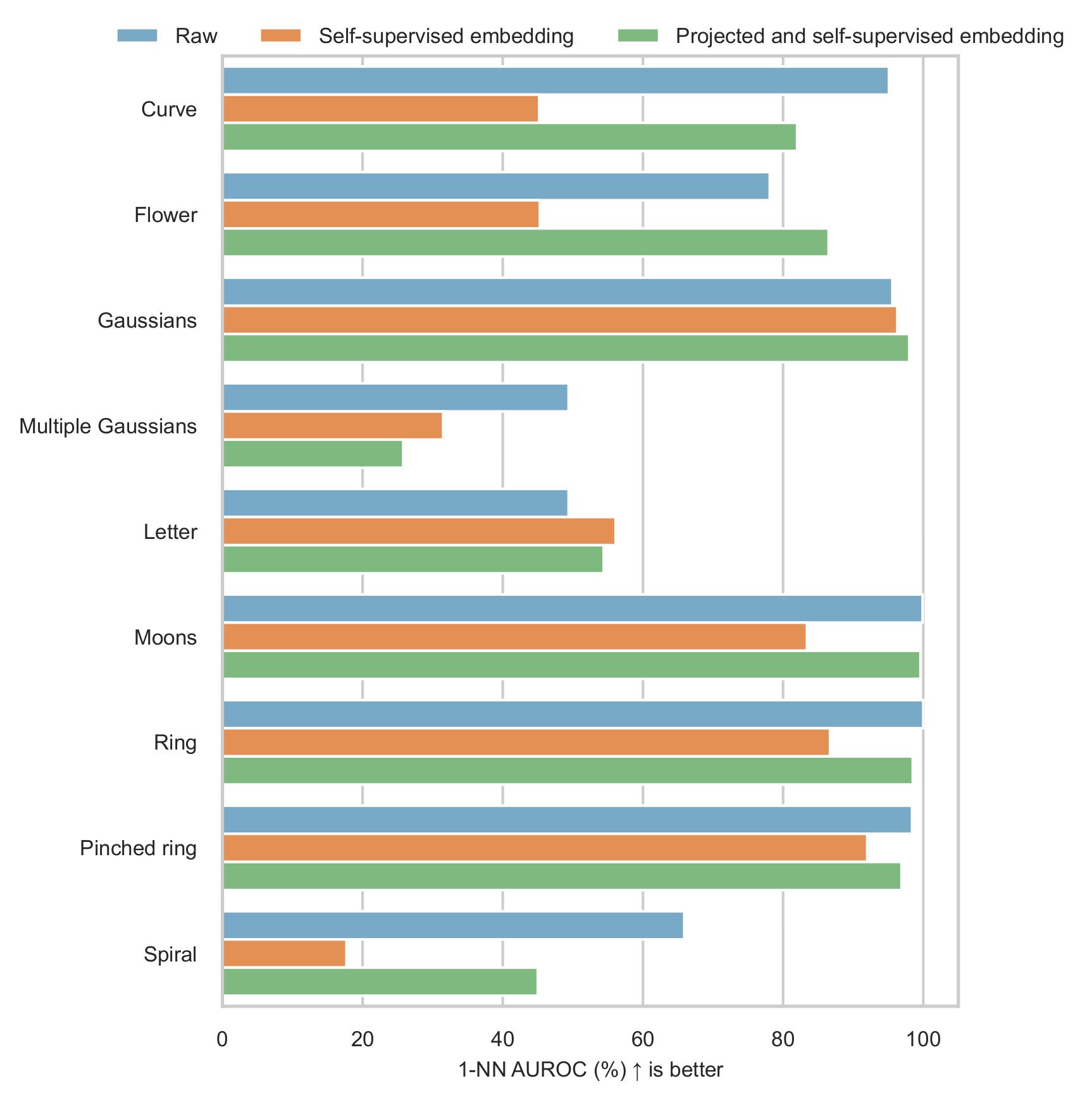}
    \caption{Nearest neighbour performance on the toy datasets. The raw embedding (blue) is the best in almost all instances. However, the self-supervision embeddings (orange) improve when projecting to a lower dimensional space (green).}
    \label{fig:toy_nn}
\end{figure}

We project the embeddings extracted from the ResNets to a lower dimensional space using the residual eigenvectors from the training data to verify whether the curse of dimensionality affects performance. We conduct this projection because the residual norm method outperforms $k$-NN in the self-supervised experiments. Therefore, we hypothesise that projecting to a smaller space should reduce the distracting influence of the primary principal components. Consequently, running shallow detectors in this new space should garner improvements. We discard half of the directions for the toy experiments to form 64-dimensional embeddings. The anomaly detectors perform better in this new space (Figure \ref{fig:toy_nn}), corroborating the view that the neural network embeddings introduce irrelevant directions.

We can also use the toy scenarios to attempt to understand the behaviour of the detectors such as OCSVM. Our experiments suggest OCSVM fails when anomalies lie in the centre of the normal data. For example, the AUROC for OCSVM trained on the raw ring data signalled random performance at 50\%, whereas $k$-NN could detect the anomalies perfectly.
\subsection{Analysing ODDS embeddings}
\label{sec4:oddsanalysis}

\begin{figure}[!htbp]
    \centering
    \begin{subfigure}[b]{0.8\linewidth}
         \centering
         \includegraphics[width=\textwidth]{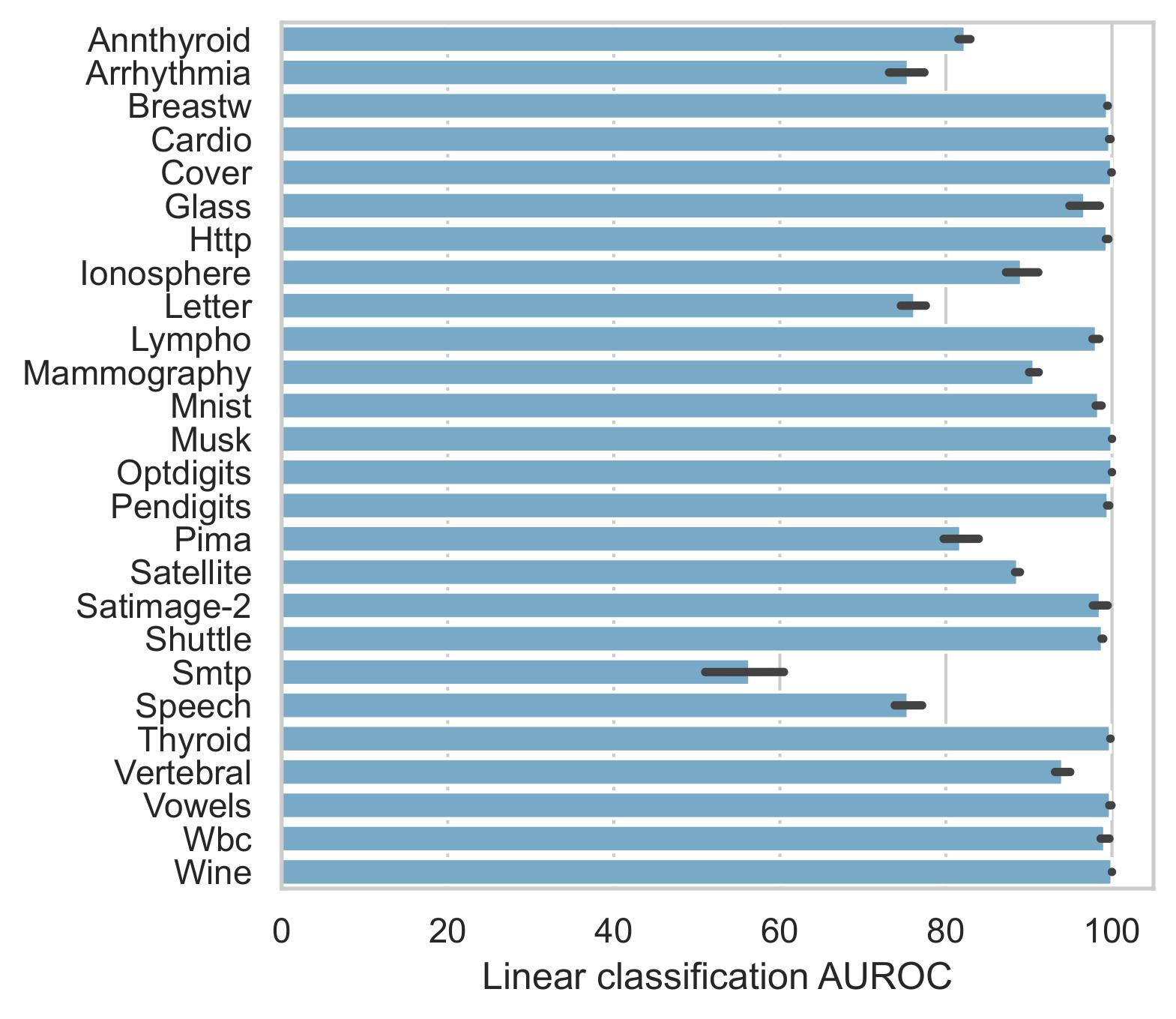}
         \caption{Classification results on the raw embeddings.}
         \label{fig:linear_classification_raw}
     \end{subfigure}
     \begin{subfigure}[b]{0.8\linewidth}
        \centering
         \includegraphics[width=\textwidth]{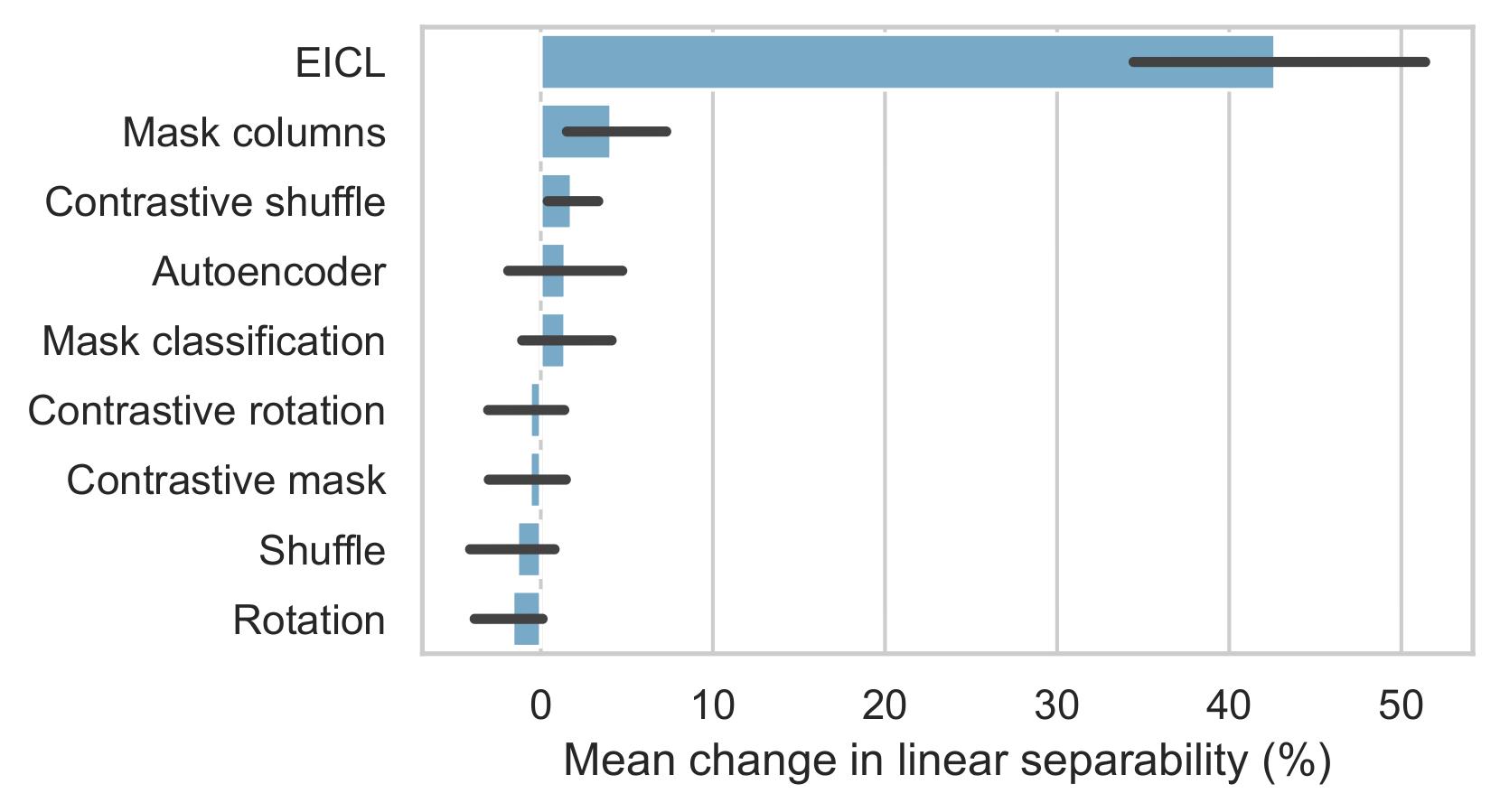}
         \caption{Differences between linear classification performance on the raw embeddings compared to the self-supervised embeddings, aggregated across the ODDS datasets. Changes greater than 0 mean the self-supervision embedding reduced separability.}
         \label{fig:linear_classification_diff}
     \end{subfigure}
    \caption{Supervised linear classification results (normal versus anomaly) on raw data (a) and supervised classification comparisons against the self-supervised embeddings (b).}
    \label{fig:classification}
\end{figure}

We now proceed to run ablations on ODDS. Previous studies have shown that supervised classification performance correlates highly with out-of-distribution detection performance \cite{vaze2022ood}. Therefore, we train linear classifiers on the self-supervised and original representation and compare classification performance. If there is a drop in performance on the self-supervised embeddings, the results would suggest the neural networks transform the data in a way that mixes anomalies with the normal samples. We could consequently attribute the poor self-supervised performance to this mixing rather than the presence of irrelevant directions.

Figure \ref{fig:linear_classification_raw} illustrates classification scores on the raw data. Most datasets are almost perfectly linearly separable in this embedding space, indicating that anomaly detectors should perform well. Figure \ref{fig:linear_classification_diff} depicts the mean difference between the raw and self-supervised classification performances. Except for EICL, the differences between linear classification performance on the raw embeddings and the self-supervised embeddings are close to zero. These trends suggest the self-supervised embeddings retain reasonable separability between the normal data and anomalies. We can rule out the mixing effect and conclude that self-supervision generally does not affect the separability between the two classes.

\clearpage
\begin{figure}[!htbp]
    \centering
    \includegraphics[width=0.8\linewidth]{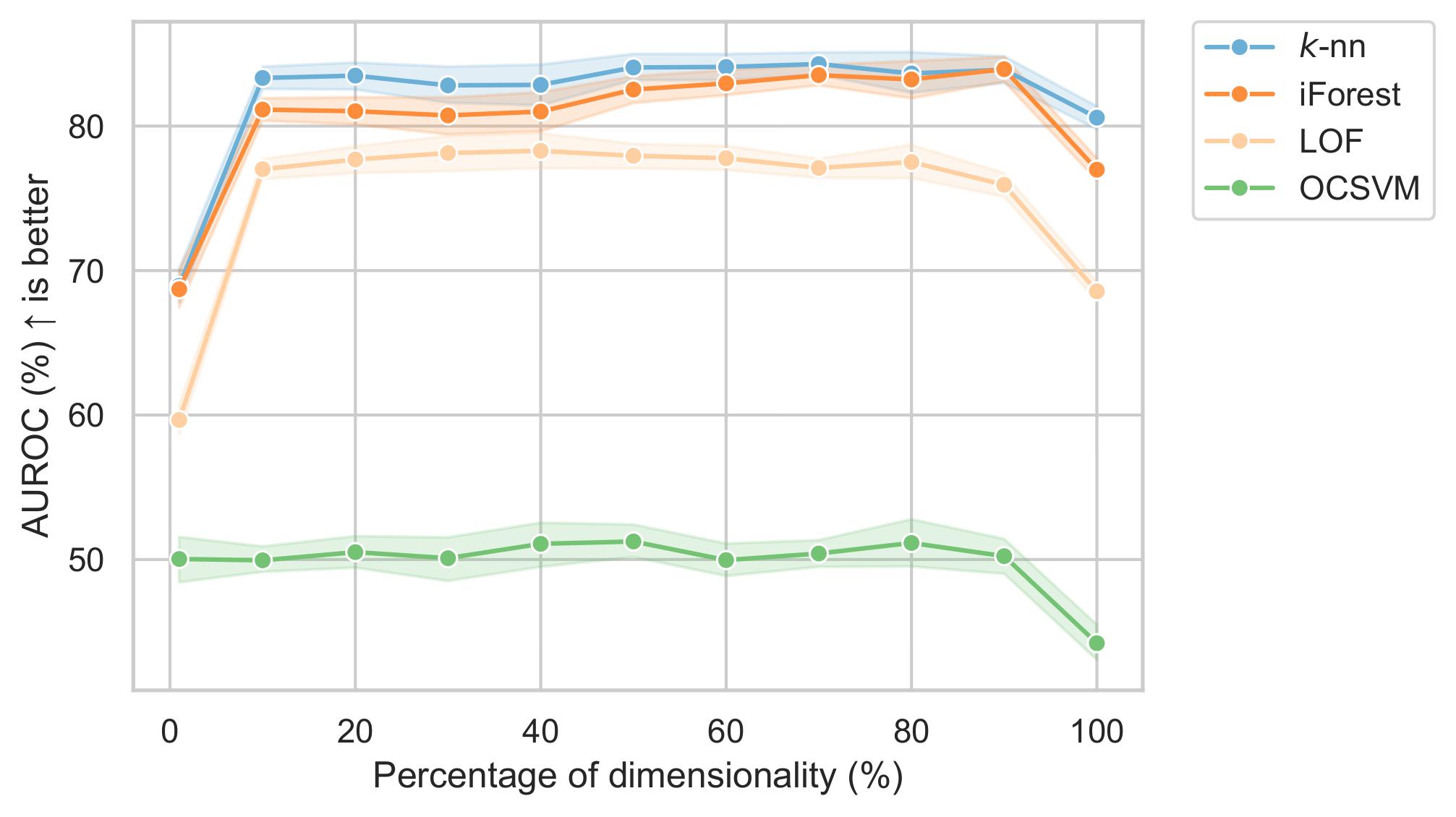}
    \caption{Ablation study showing how shallow detector results vary with subspace dimensionality.}
    \label{fig:subspace}
\end{figure}
We now investigate the residual space of the embeddings by extending the toy dataset analyses to ODDS. We take the smallest eigenvalues (from 1\% to 90\% in 10\% increments) to project the neural network embeddings to their residual representations. We proceed to re-run the shallow anomaly detectors in the new space. Figure \ref{fig:subspace} shows the results. We aggregate both ResNet and FT-Transformer scores as observed similar behaviour across the two architectures. Reducing the dimensionality indeed boosts performance. 

{On all of the shallow detectors, using the entire representation space (100\% dimensionality in Figure \ref{fig:subspace}) results in lower AUROCs than using a subset. Throwing away the top 10\% of principal components garners most improvements, although performance generally remains stable when discarding more components - up to the top 90\%.

This observation aligns with previous findings that show residual directions capture information important for out-of-distribution detection \cite{kamoi2020mahalanobis}. The magnitude of normal data is minute in this space, which is not necessarily the case for anomalies. Based on these results, we do not need complete neural network representations to perform anomaly detection. A subset suffices. 

\subsection{Synthetic anomalies}
\label{sec4_synthetic}

Anomaly detection depends on two factors: the nature of the normal data and the nature of anomalies. Both classes can originate from complex, irregular distributions. These aspects make it difficult to pinpoint the causes of results on ODDS and other curated datasets. We attempt to disentangle these factors by analysing performance on synthetic anomalies. The anomalies curated in ODDS are a composite of these types. We calculated the correlation between the ODDS and the synthetic anomaly scores and found that the datasets exhibited correlations between multiple synthetic categories, highlighting the complex qualities of the anomalies. For example, when analysing the raw data representations, $k$-NN on the curated \textit{Letter} anomalies correlates strongly with local ($\rho = 0.84$), global ($\rho =0.49 $), and dependency ($\rho = 0.94$) anomaly scores.

\begin{figure}[h!]
    \centering
    \begin{subfigure}[b]{0.45\linewidth}
         \centering
         \includegraphics[width=\textwidth]{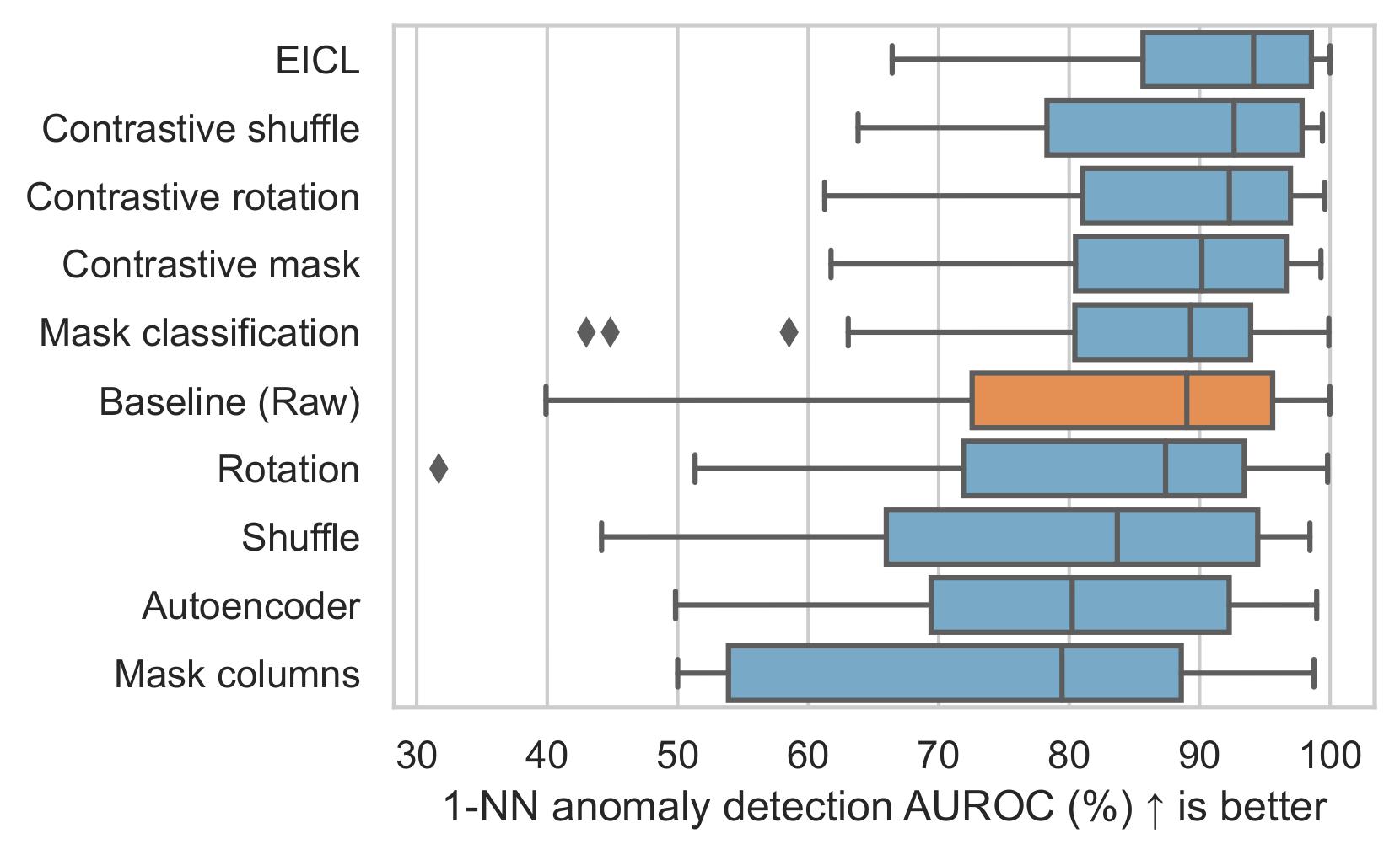}
         \caption{Local anomalies ($\alpha = 2$).}
         \label{fig:synthetic_local}
     \end{subfigure}
      \hfill
     \begin{subfigure}[b]{0.45\linewidth}
     \centering
     \includegraphics[width=\textwidth]{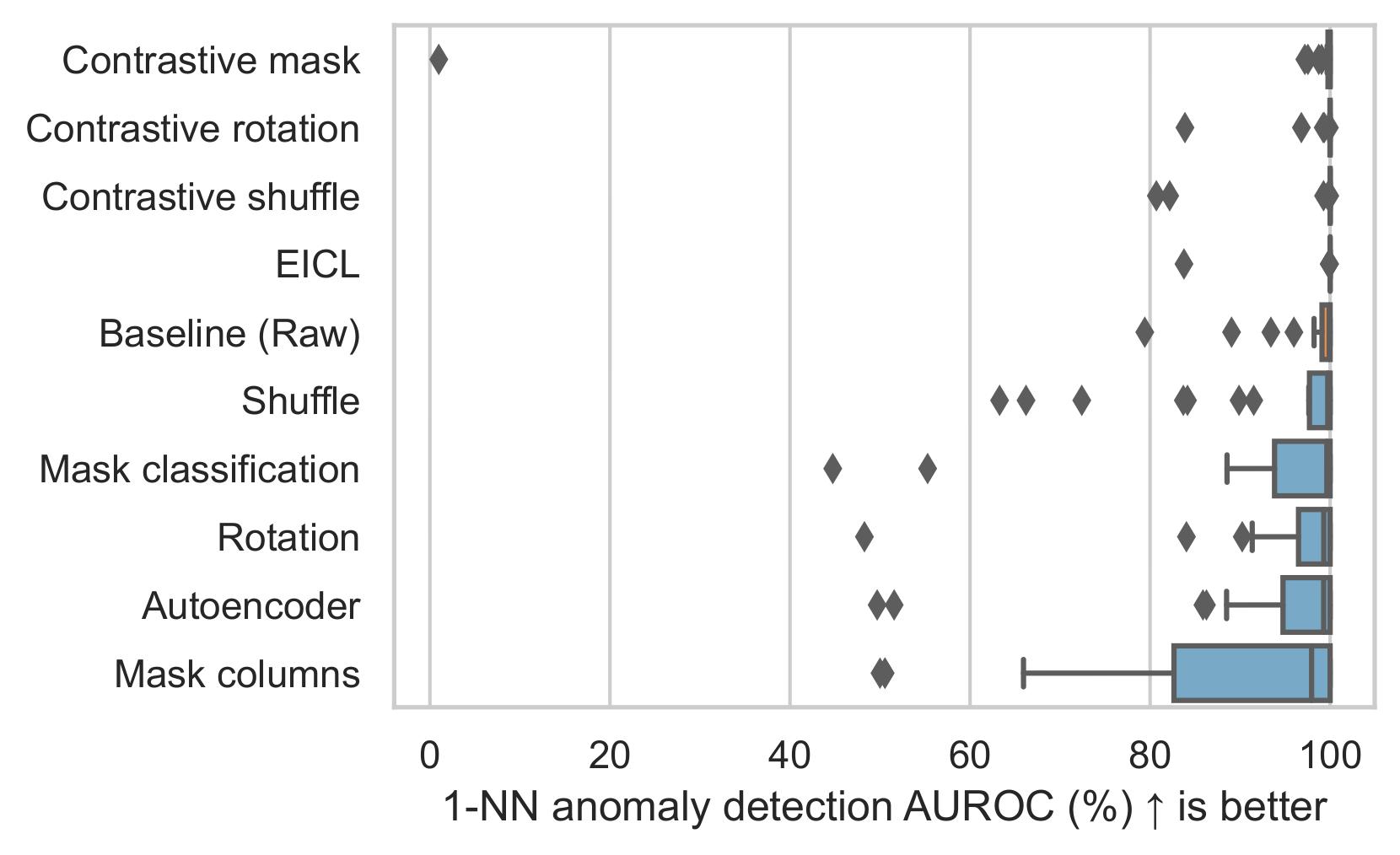}
     \caption{Cluster anomalies ($\beta$ = 2).}
     \label{fig:synthetic_cluster}
     \end{subfigure}
     \hfill
     \begin{subfigure}[b]{0.45\linewidth}
        \centering
         \includegraphics[width=\textwidth]{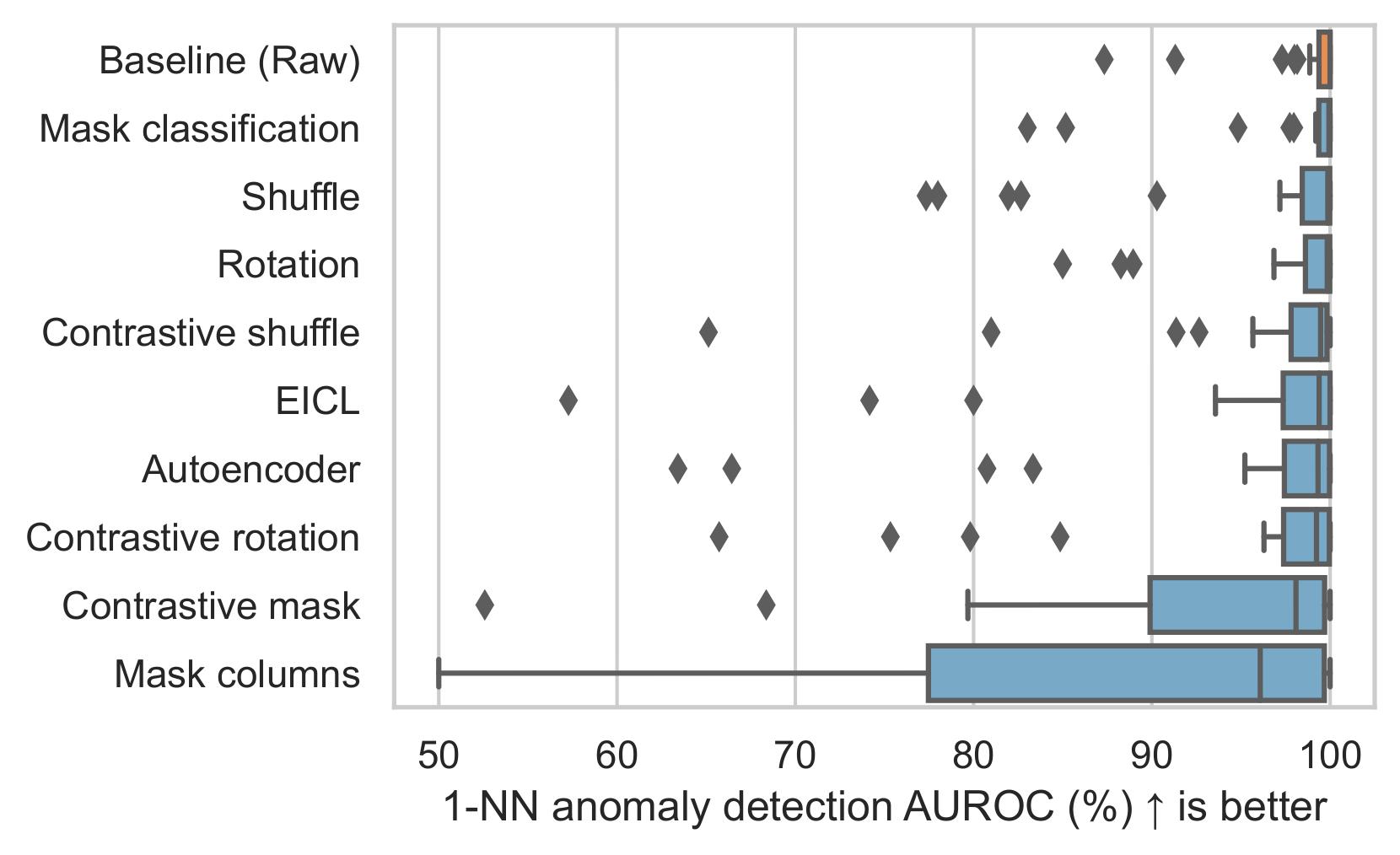}
         \caption{Global anomalies ($\delta = 0.01$).}
         \label{fig:synthetic_global}
     \end{subfigure}
     \hfill
     \begin{subfigure}[b]{0.45\linewidth}
        \centering
         \includegraphics[width=\textwidth]{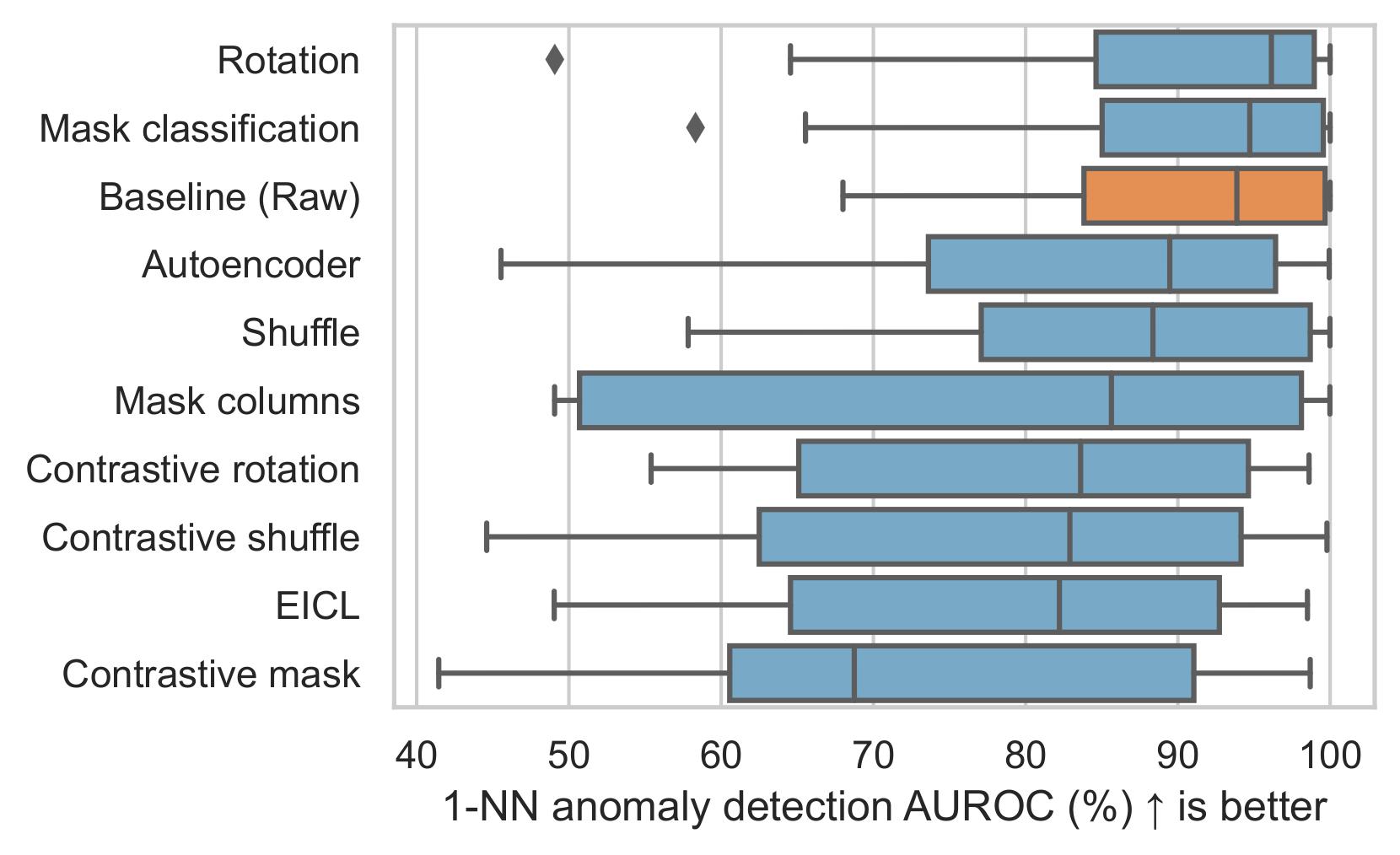}
         \caption{Dependency anomalies.}
         \label{fig:synthetic_dependency}
     \end{subfigure}
    \caption{Bar plots comparing synthetic anomaly results across the representations.}
    \label{fig:synthetic}
\end{figure}

Figures \ref{fig:synthetic_local} to \ref{fig:synthetic_dependency} show the results across the four synthetic types. We show comparisons using $k$-NN as we found similar behaviours across the detectors. The contrastive objectives outperform the baseline in the local (Figure \ref{fig:synthetic_local}) and cluster anomaly (Figure \ref{fig:synthetic_cluster}) scenarios. This result suggests contrastive tasks are better at discerning differences at a local neighbourhood level. 

No self-supervised approach beats the baseline when faced with global anomalies (Figure \ref{fig:synthetic_global}). This result contributes to the idea that self-supervised representations introduce irrelevant directions. Since the global anomalies scatter across the representation space, these additional directions mask the meaningful distances between the anomalies and normal points. As a result, methods like $k$-NN become less effective. In addition, the ranking of the self-supervised tasks aligns most closely with their rankings on ODDS (Figure \ref{fig:overall_cd}), which potentially highlights the overall properties of the ODDS datasets.

For the dependency anomalies, rotation and mask classification surpass the baseline (Figure \ref{fig:synthetic_dependency}). Conversely, contrastive tasks perform the worst. Using a rotation or mask classification pretext task could help promote the intrinsic property that tabular data are non-invariant, which may help identify this type of anomaly.

\subsection{Architectural choices for self-supervision}
\label{sec4_architecture}

\begin{figure}[!htbp]
    \centering
    \begin{subfigure}[b]{0.35\linewidth}
         \centering
         \includegraphics[width=\textwidth]{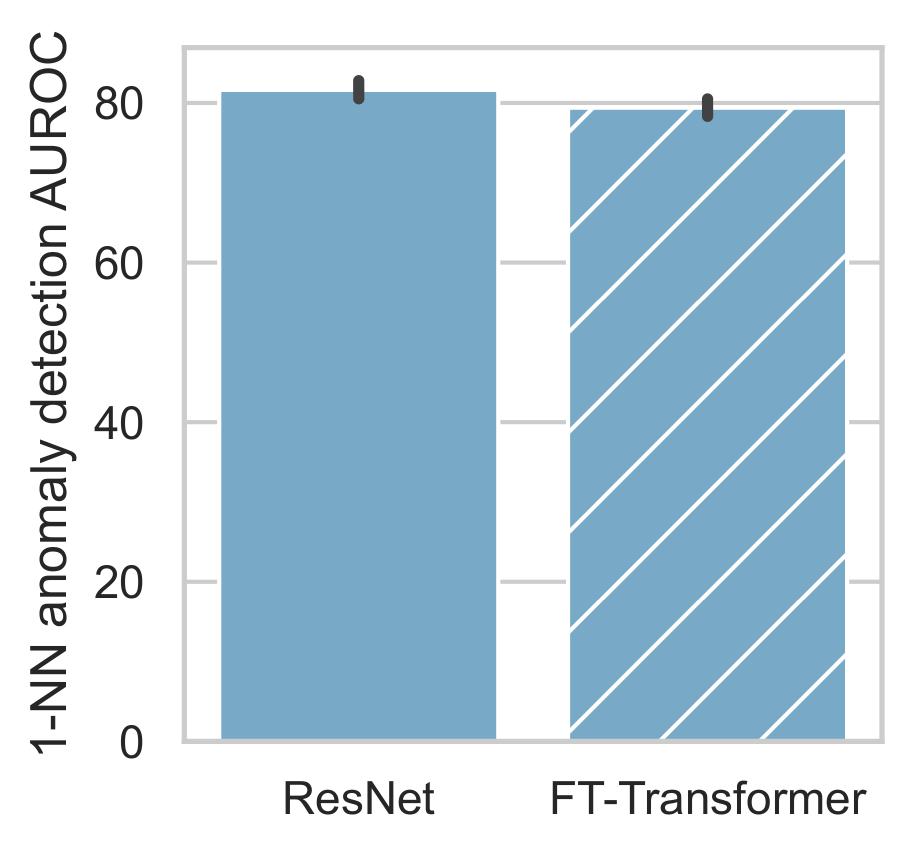}
         \caption{Architectural results.}
         \label{fig:choice_architecture}
     \end{subfigure}
     \quad
     \begin{subfigure}[b]{0.35\linewidth}
        \centering
         \includegraphics[width=\textwidth]{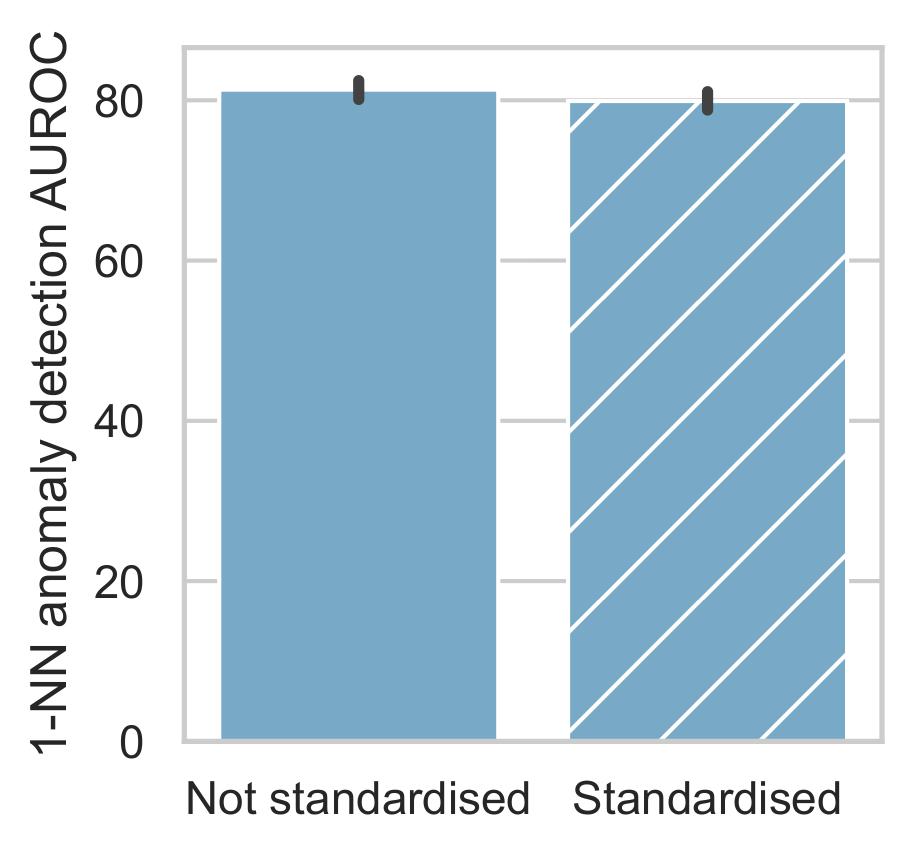}
         \caption{Standardisation results.}
         \label{fig:choice_standardisation}
     \end{subfigure}
     \quad
         \begin{subfigure}[b]{0.35\linewidth}
         \centering
         \includegraphics[width=\textwidth]{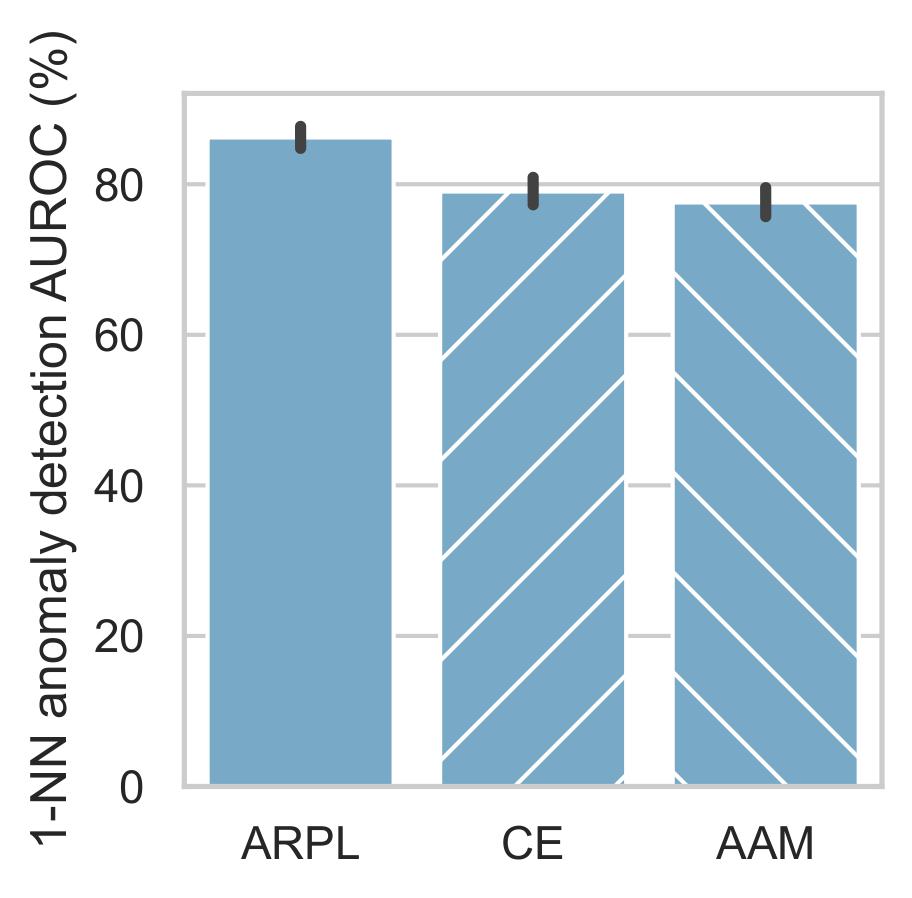}
         \caption{Classification losses.}
         \label{fig:choice_classification}
     \end{subfigure}
     \quad
     \begin{subfigure}[b]{0.35\linewidth}
        \centering
         \includegraphics[width=\textwidth]{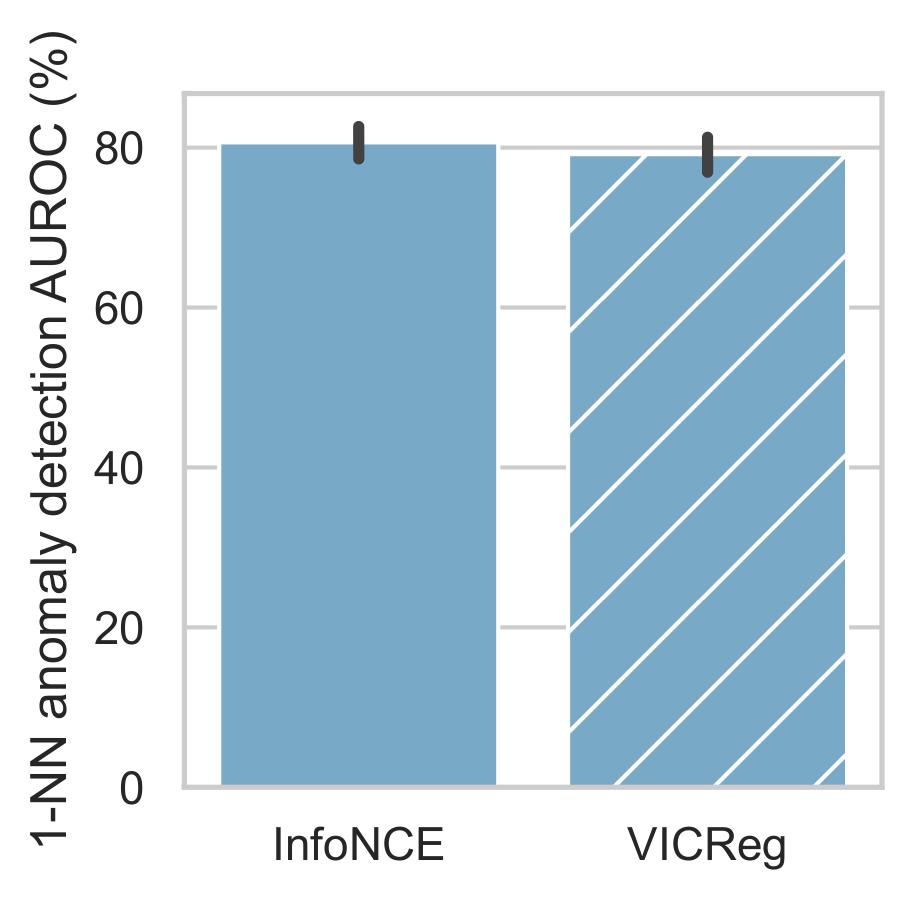}
         \caption{Contrastive losses.}
         \label{fig:choice_contrastive}
     \end{subfigure}
    \caption{Comparisons of how architecture and losses affect performance on the self-supervised embeddings.}
    \label{fig:choice}
\end{figure}

We analyse the effects of architectures and loss functions on performance to provide starting points for improving deep learning methods for tabular anomaly detection. We illustrate the results using $k$-NN as we observe similar behaviours across detectors.

\textbf{ResNets outperform transformers}. Our experiments indicate ResNets are a better choice than FT-Transformer (Figure \ref{fig:choice_architecture}). This result may be due to transformers needing more training data during the learning phase \cite{dosovitsky2021image} - the ODDS datasets are relatively small. 

\textbf{Standardisation is not necessary}. Standardising data before training neural networks does not offer much benefit (Figure \ref{fig:choice_standardisation}).

\textbf{ARPL is a better choice for classification-type losses}. ARPL significantly outperforms cross-entropy and AAM when training classification-type tasks (Figure \ref{fig:choice_classification}). Specialised losses like ARPL might represent ``other'' spaces better in the context of smaller datasets.

\textbf{InfoNCE is better than VICReg for contrastive-type losses}. This result (Figure \ref{fig:choice_contrastive}) may be due to the intricacies of VICReg, which requires balancing three components (pair similarity, variance and covariance). 

\subsection{Benchmarking unsupervised anomaly detection}
\label{sec4:benchmark}

Finally, we compare the performance of each of the detectors overall to see how well they perform in one-class settings. We aggregate results across the baseline and self-supervised embeddings to provide a more generalised understanding of detector behaviour. 

\begin{figure}[h!]
    \centering
    \includegraphics[width=0.8\linewidth]{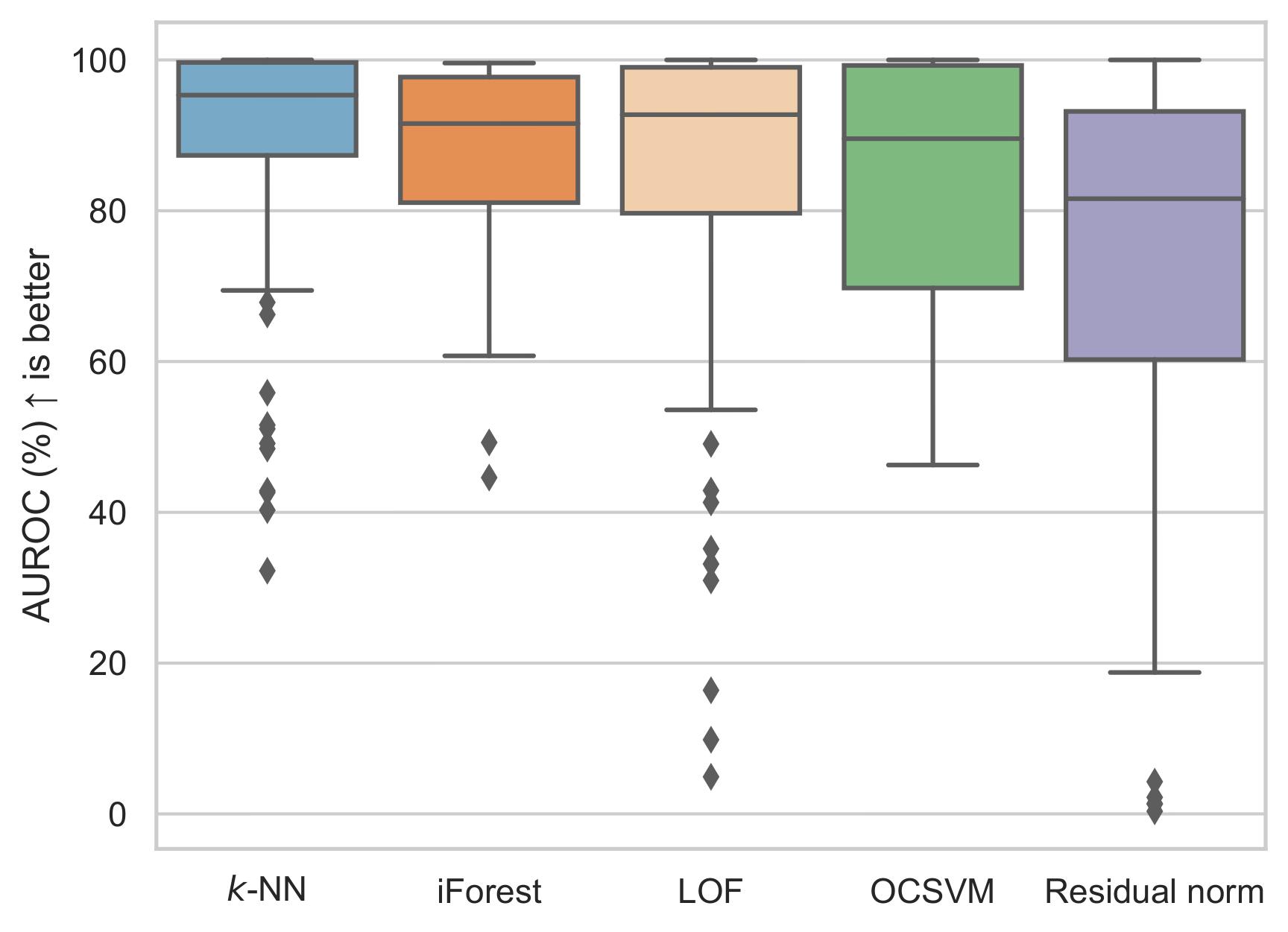}
    \caption{Box plot comparing detector performance on the raw and standardised data. The results include all hyperparameter variations where available.}
    \label{fig:overall_boxplot}
\end{figure}

\begin{figure}[h!]
    \centering
    \includegraphics[width=0.8\linewidth]{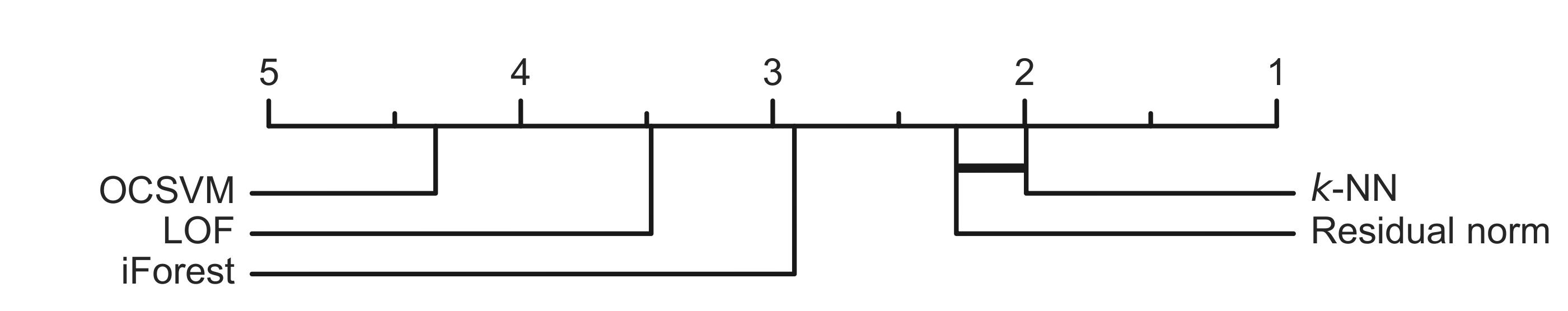}
    \caption{Critical difference diagram ranking the different detectors. The dark lines between different detectors indicate a statistical difference ($p < 0.05$) in results when running pairwise comparison tests.}
    \label{fig:overall_cd}
\end{figure}

Figures \ref{fig:overall_boxplot} and \ref{fig:overall_cd} summarise the overall performances of each anomaly detector on ODDS. Even with the inclusion of self-supervised representations, $k$-NN performs best. Our findings align with other works highlighting $k$-NN as a performant anomaly detector \cite{gu2019knn, reiss2021panda, sun2022knn, sehwag2021ssd}. However, apart from $k$-NN and residual norm, Figure \ref{fig:overall_cd} shows no significant statistical differences between the detectors, suggesting the detectors make similar classification decisions. $k$-NN might be a sensible starting point that does not make strong assumptions about the normal distribution. Nonetheless, the choice of underlying representation should take precedence over the detector when designing anomaly detection systems.

\subsubsection{Hyperparameter ablations}
We now examine the sensitivity of the detectors to changes in hyperparameters. These experiments were conducted directly on the raw ODDS data only to understand detector performance in an optimal representation space. By doing so, these results enable a better understanding of the detectors' inductive biases and why they may deteriorate in suboptimal self-supervised representations. 

\begin{figure}[!htbp]
    \centering
    \includegraphics[width=0.6\linewidth]{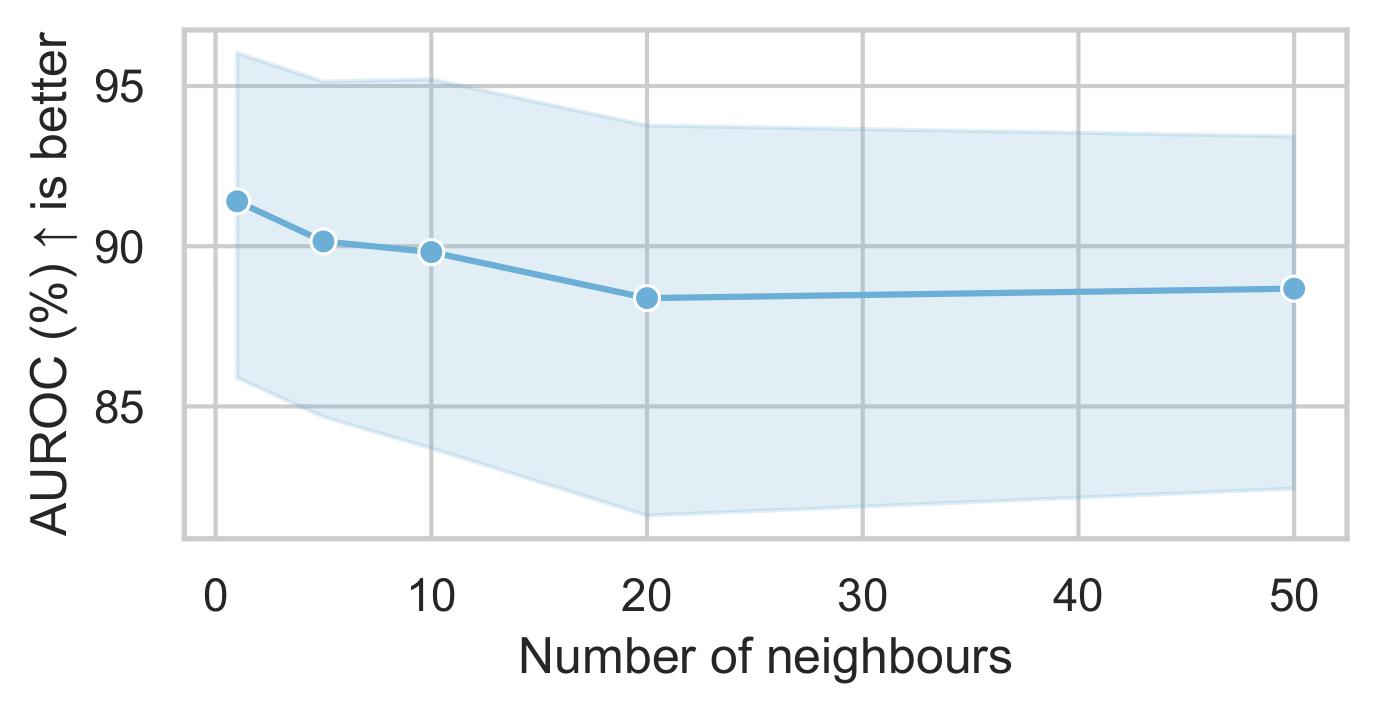}
    \caption{Line plot showing how $k$-NN varies with the change in the number of nearest neighbours, aggregated across the ODDS datasets, with 95\% confidence intervals.}
    \label{fig:knn}
\end{figure}

\textbf{$k$-NN}: Figure \ref{fig:knn} shows performance remains relatively stable to changes in $k$, suggesting the choice of this hyperparameter is trivial. As $k$-NN considers global relationships, this result indicates that anomalies already lie in distinct regions separate from the normal raw data.

\clearpage
\begin{figure}[!htbp]
    \centering
    \includegraphics[width=0.6\linewidth]{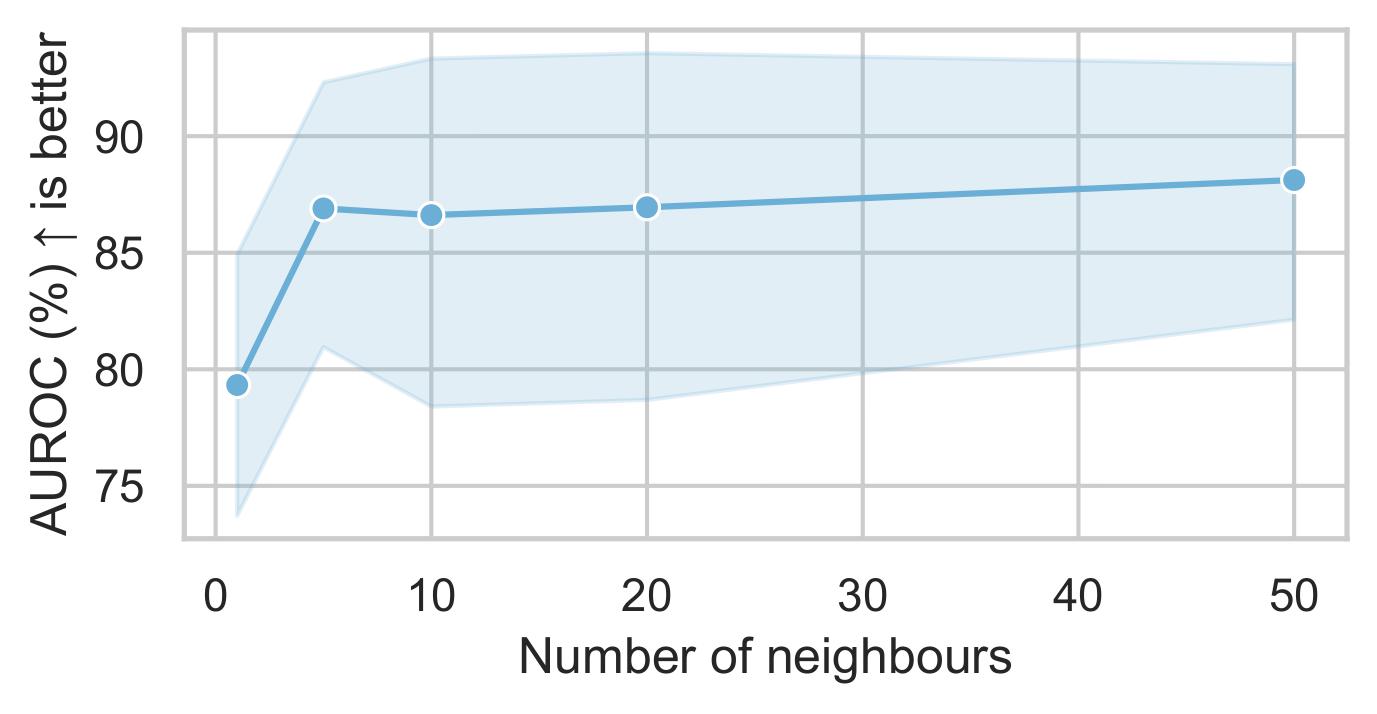}
    \caption{Line plot showing how LOF varies with the change in the number of nearest neighbours, aggregated across the ODDS dataset, with 95\% confidence intervals.}
    \label{fig:lof}
\end{figure}

\textbf{LOF}: Figure \ref{fig:lof} illustrates how LOF performance changes with $k$. Although LOF and $k$-NN consider points in a neighbourhood, LOF is more sensitive to the number of neighbours (as evidenced by the increase in performance when $k$ = 1 and $k$ = 5 for LOF). However, it is unclear how to choose a value of $k$ so that LOF is competitive with the other detectors in the one-class setting.

\begin{figure}[!htbp]
    \centering
    \includegraphics[width=0.6\linewidth]{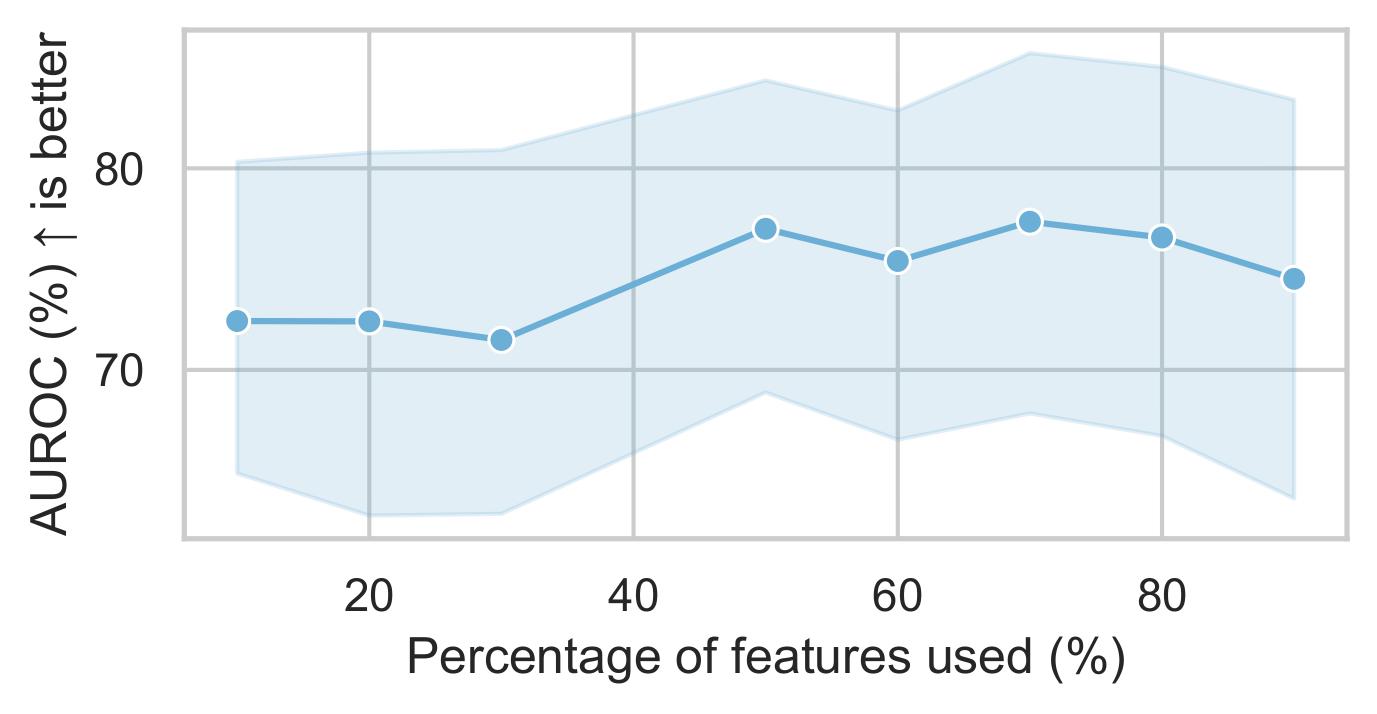}
    \caption{Line plot showing how residual norm varies with the change in residual dimensionality, aggregated across the ODDS dataset, with 95\% confidence intervals.}
    \label{fig:norm}
\end{figure}

\textbf{Residual norms}: Figure \ref{fig:norm} shows how performance varies with the percentage of attributes used. There are no notable trends, although performance remains better than random, even with a small subset (10\%) of features. The number of relevant attributes in the original representation space is dataset-dependent as ODDS contains datasets from differing tasks. It is unclear how to choose the number of features to maximise the performance of residual norms in the original dataset space.
\clearpage
\subsubsection{Corrupted input data}

\begin{figure}[!htbp]
    \centering
    \begin{subfigure}[b]{0.45\linewidth}
         \centering
         \includegraphics[width=\textwidth]{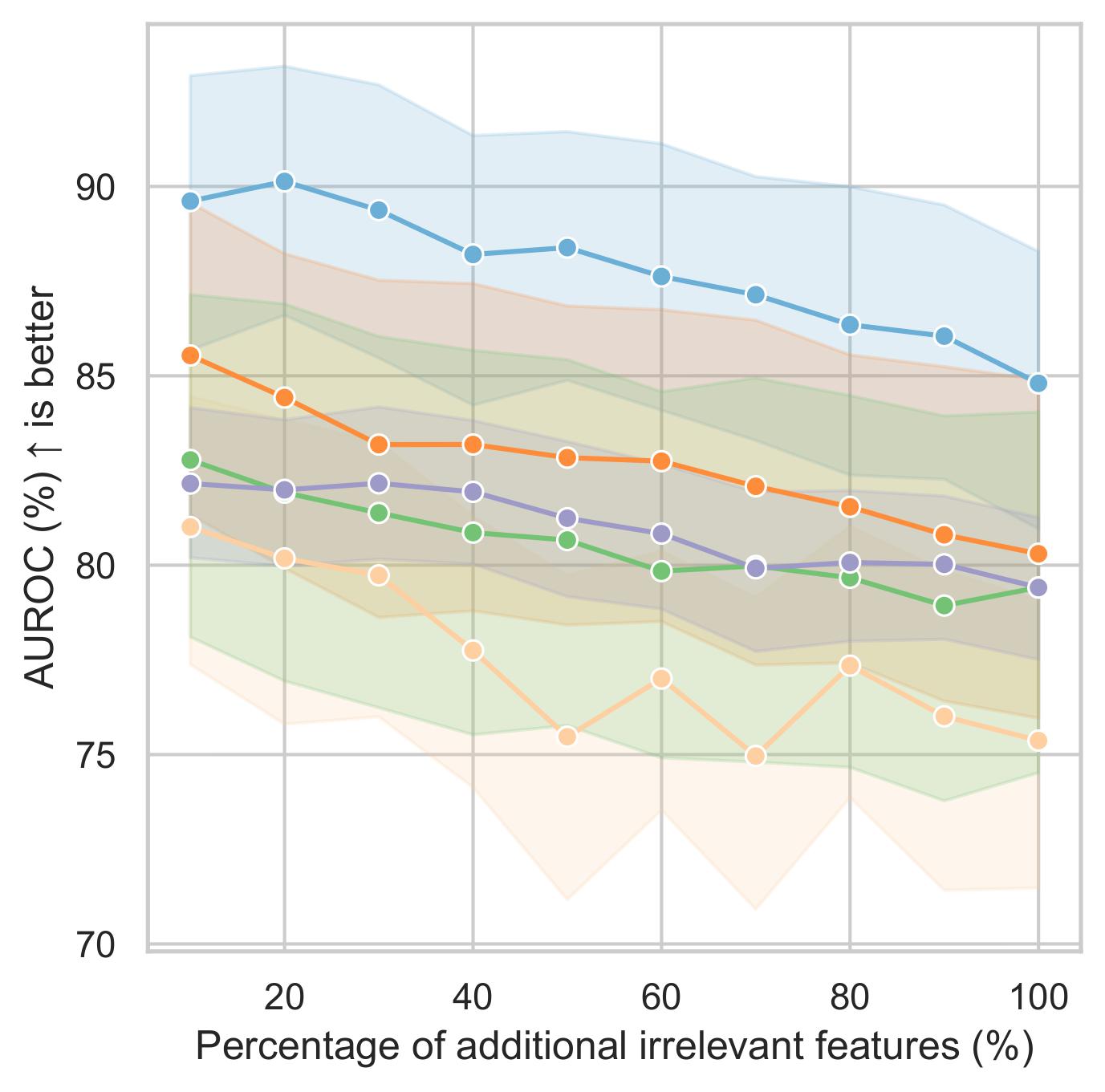}
         \caption{Additional features.}
         \label{fig:corrupt_add}
     \end{subfigure}
     \quad
     \begin{subfigure}[b]{0.45\linewidth}
        \centering
         \includegraphics[width=\textwidth]{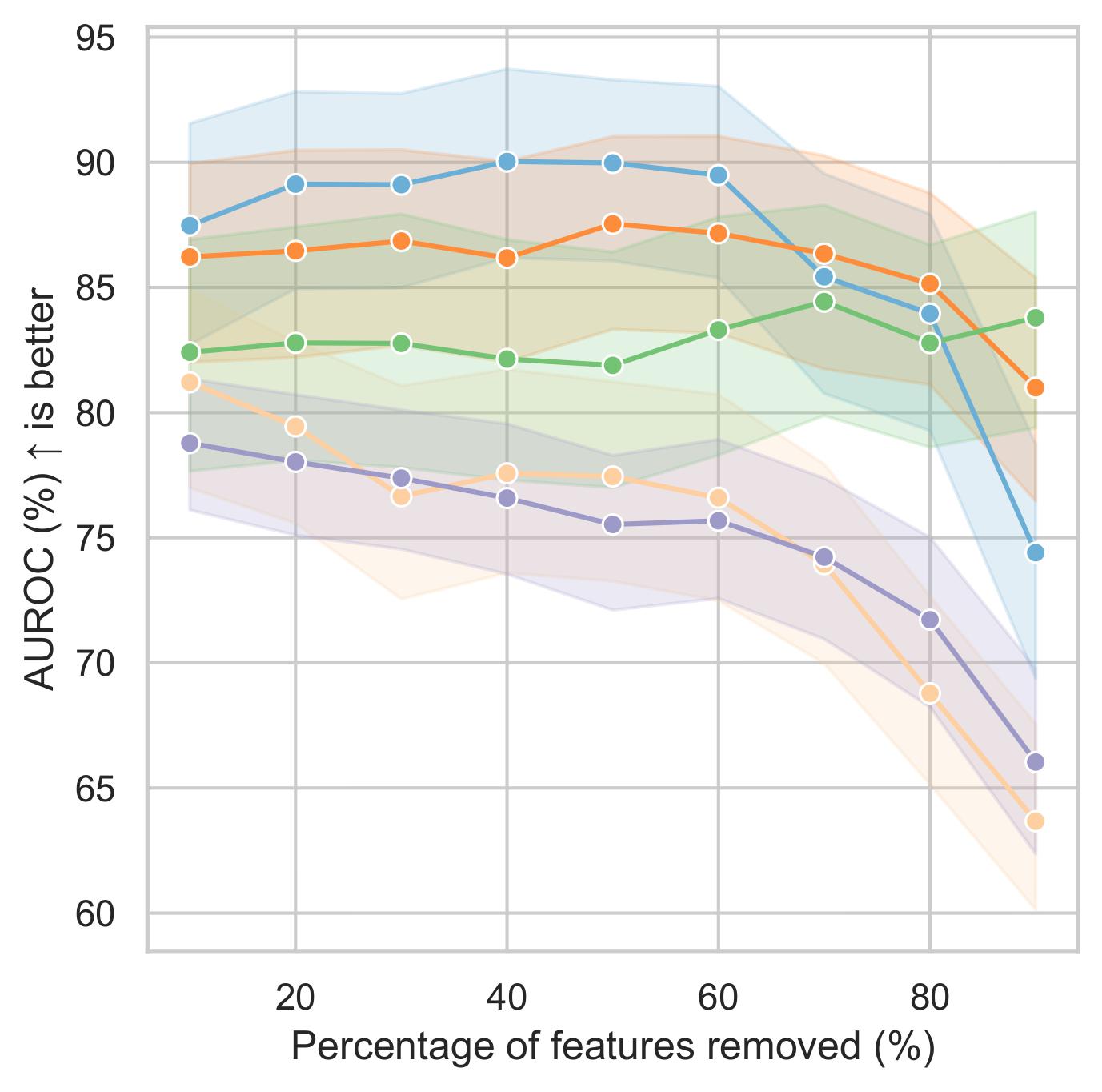}
         \caption{Removing features.}
         \label{fig:corrupt_remove}
     \end{subfigure}
    \begin{subfigure}[b]{0.45\linewidth}
         \centering
         \includegraphics[width=\textwidth]{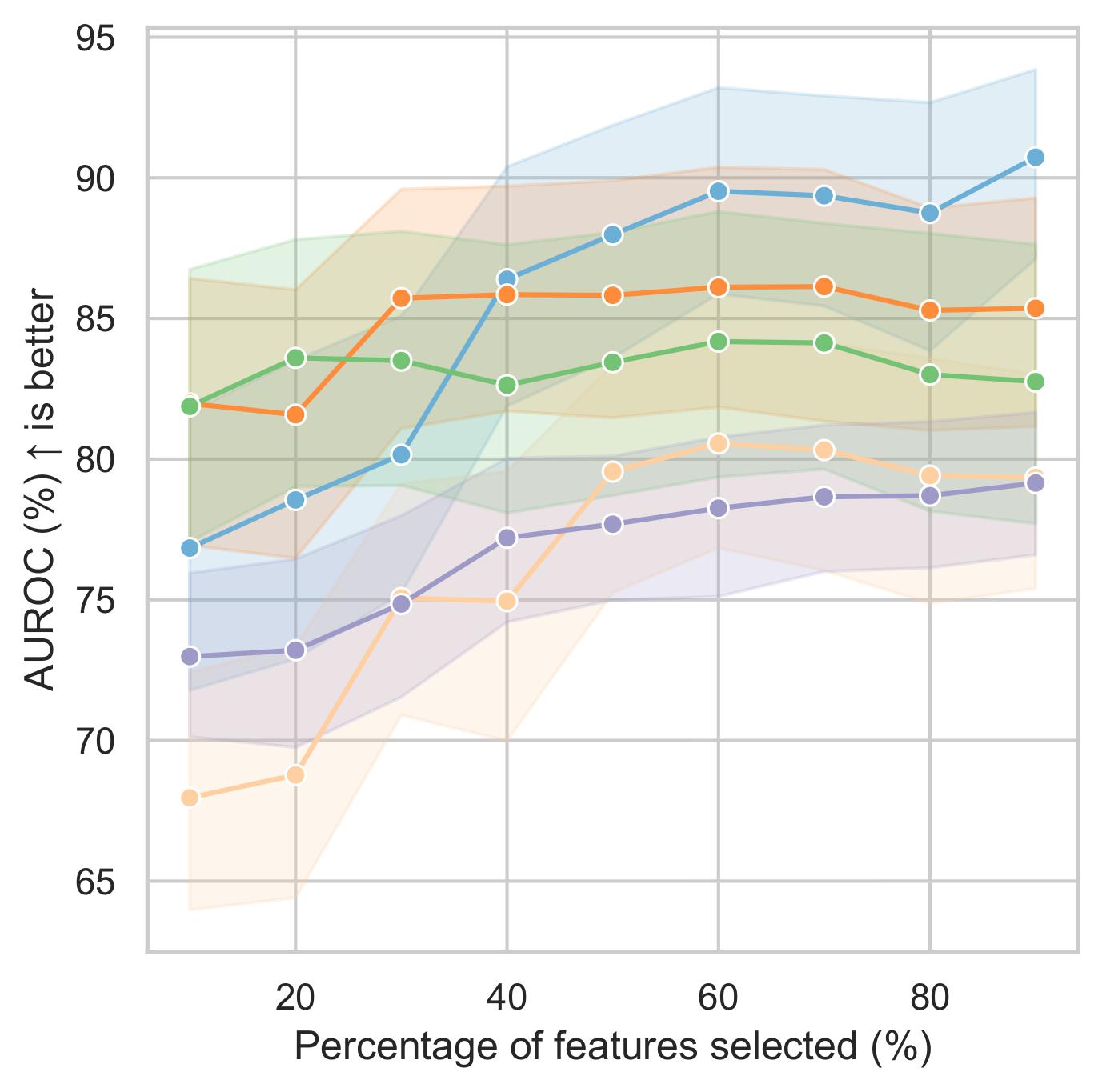}
         \caption{Selecting features.}
         \label{fig:corrupt_select}
     \end{subfigure}
     \quad
     \begin{subfigure}[b]{0.45\linewidth}
        \centering
         \includegraphics[width=\textwidth]{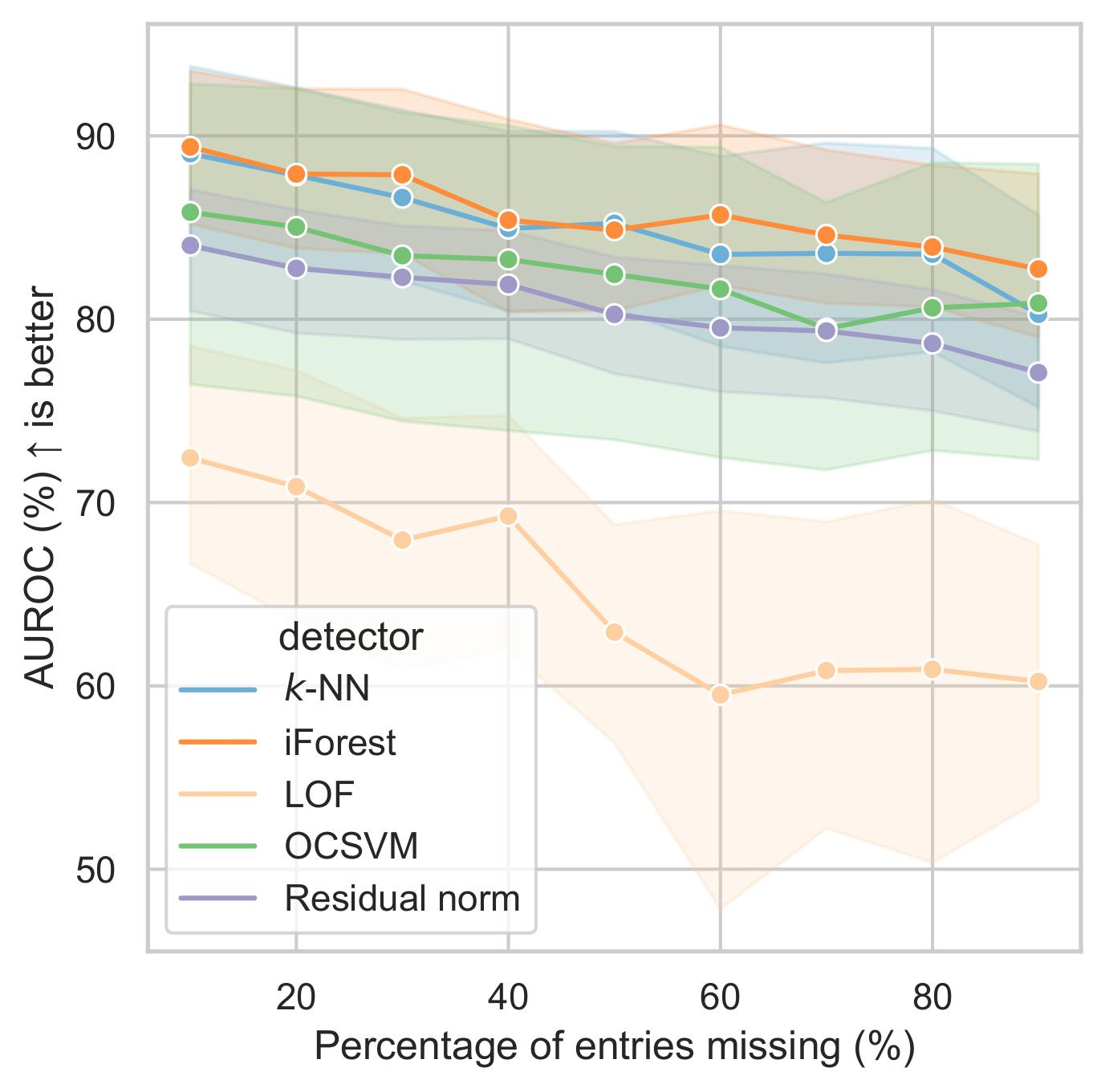}
         \caption{Missing values.}
         \label{fig:corrupt_missing}
     \end{subfigure}
    \caption{Ablations showing how detector performance varies with changing levels of corrupt data.}
    \label{fig:corrupt}
\end{figure}

\textbf{Adding uninformative features}: All detectors are sensitive to irrelevant features (Figure \ref{fig:corrupt_add}). Although residual norms do not achieve the highest performance, it is more stable under increasing noise levels. This result may be due to the residuals capturing the most meaningful directions of the data. In contrast, $k$-NN performance declines the most. 

\textbf{Removing and selecting important features}: Overall, performance plateaus at around 50\% of attributes, suggesting half of the raw features are irrelevant for anomaly detection. iForest and OCSVM are the most stable under varying subsets of features (Figures \ref{fig:corrupt_remove} and \ref{fig:corrupt_select}). 

\textbf{Missing values}: Most detectors exhibit a slight decline in AUROC with increasing proportions of missing values (Figure \ref{fig:corrupt_missing}). LOF is the exception, as performance drops significantly.

Overall, the results indicate $k$-NN is the best-performing detector when faced with clean and relevant features. However, the relative ranking of detectors changes in the presence of corrupted input data. As observed in our self-supervised results (Section \ref{sec4:oddsanalysis}), residual norms might be better at filtering out noisy directions. Furthermore, when there are fewer relevant features, iForest may be a better choice.  

\section{Conclusion}\label{sec5}

\subsection{Limitations and future work}
We limited our experiments to the ODDS, which is not necessarily representative of all tabular anomaly datasets. Several datasets underwent preprocessing during the curation of ODDS, which could affect results. For example, the values in \textit{HTTP} were log-transformed. In addition, the datasets are relatively small. As neural networks (particularly transformers) benefit from large amounts of data \cite{dosovitsky2021image}, it is unclear if self-supervision would be more advantageous in the big data case. Contrastive objectives are particularly reliant on large datasets and batch sizes \cite{oord2016cpc, chen2020simclr} Additional ablations could examine the effects of dataset size on representation quality and detection performance.

Furthermore, we isolated our analyses by extracting embeddings at the penultimate layer and running shallow anomaly detection algorithms. Although feature extraction at this stage combined with simple detectors is a popular strategy \cite{gu2019knn, reiss2021panda, sun2022knn, sehwag2021ssd}, different parts of the neural network could provide more informative features \cite{kim2020rapp}. Moreover, we chose to use shallow detectors to prioritise studying the effect of representations rather than studying the detection approach. In addition, the original implementations of ICL and GOAD evaluate anomalies using an entire neural network pipeline and use specific architectures for the tasks. Adapting these implementations for a pretext task with different architectures deviates from the original setup and could affect performance. Future work could look at extending the experiments to examine how varying pretext tasks with deep anomaly detection can yield better results \cite{ruff2018svdd}. 

Another direction for future work that focuses on representation quality could replace the one-class detectors with semi-supervised or supervised classifiers. We decided to concentrate on one-class detectors to align with the anomaly detection field \cite{reiss2022ad, scholkopf1999ocsvm, chandola2009ad, ruff2021ad}. However, anomalies can manifest in different ways, and it could be challenging for an unsupervised detector to capture the relevant features for a specific task in practice. Incorporating prior knowledge about anomalies through weak or semi-supervised detection approaches could improve detection \cite{ruff2020ad}.

In addition, studies focusing on improving deep tabular anomaly detectors could also start examining regularisation strategies. Our experiments suggest neural networks add irrelevant features, hence regularisation during the training process could help to control this behaviour.

\subsection{Summary}
We trained multiple neural networks on various self-supervised pretext tasks to learn new representations for ODDS, a series of tabular anomaly detection datasets. We ran a suite of shallow anomaly detectors on the new embeddings and compared the results to the performance of the original data. None of the self-supervised representations outperformed the raw baseline. 

We conducted ablations to try to understand this behaviour. Our empirical findings suggested that neural networks introduce irrelevant features, which degrade detector capability. As normal and anomalous data were easily distinguishable in the original tabular representations, neural networks merely stretched the data. They did not introduce any additional informative information. However, we demonstrated performance was recoverable by projecting the embeddings to a residual subspace.

As the anomalies from ODDS derive from complex distributions, we repeated the experiments on synthetic data to understand the pretext tasks' influence on detecting particular anomaly types. We showed in specific scenarios that self-supervision can be beneficial. Contrastive tasks were better at picking up localised anomalies, while classification tasks were better at identifying differences in dependency structures. 

Finally, we studied different shallow detectors by aggregating performances across the baseline and self-supervised representations. We showed that localised methods like $k$-NN and LOF worked best on ODDS but were susceptible to performance degradation with corrupted data. In contrast, iForest was more robust. Our findings provided practical insights into when one detector might be preferable to another. 

Overall, our findings complement the growing landscape of theories on why self-supervised learning works. Effective self-supervised pretext tasks learn to compress the input data when there are irrelevant features \cite{lee2021compressive, schwarz-ziv2023ssl, yu2023whitebox}. Our findings suggest current deep learning approaches do not add much benefit when the original feature space succinctly represents the normal data. This situation is often the case for tabular data, and we demonstrated this by showing performance degrades when removing features in the original space. If the feature space did not succinctly represent the normal data, we would not observe such large degradations. This setup differs from other domains. For example, pixels in images contain lots of semantically irrelevant information. Therefore, neural networks can distil information from pixels to extract useful semantic features and self-supervision is beneficial. 



\section{Data availability}
Publicly available datasets were analysed in this study. The ODDS datasets are accessible from \url{https://odds.cs.stonybrook.edu/}.
\clearpage
\begin{appendices}

\section{Additional figures}\label{secA1}

\begin{figure}[h!]
     \centering
     \begin{subfigure}[b]{0.45\linewidth}
         \centering
         \includegraphics[width=\textwidth]{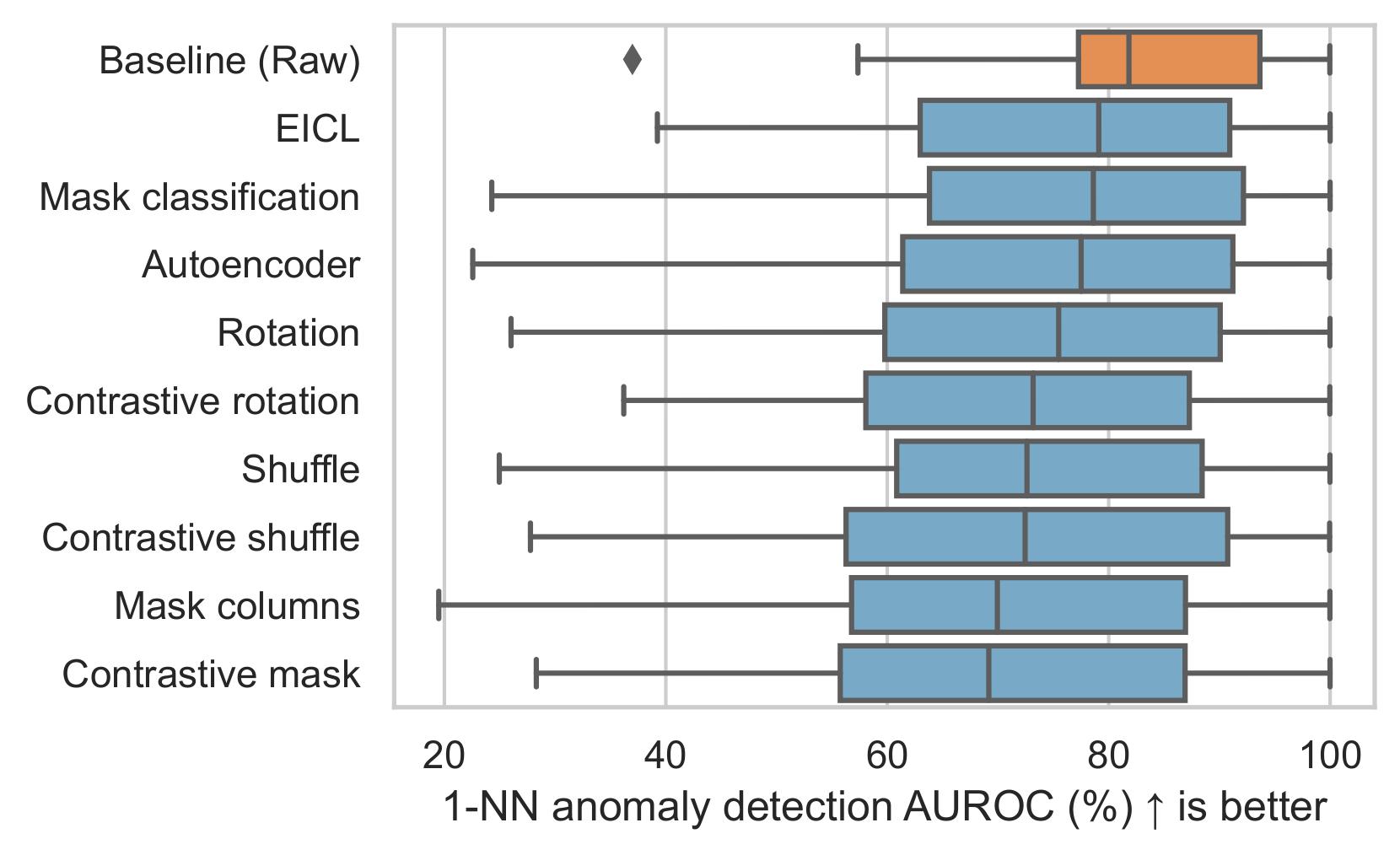}
         \caption{100\% additional features}
     \end{subfigure}
     \hfill
     \begin{subfigure}[b]{0.45\linewidth}
         \centering
         \includegraphics[width=\textwidth]{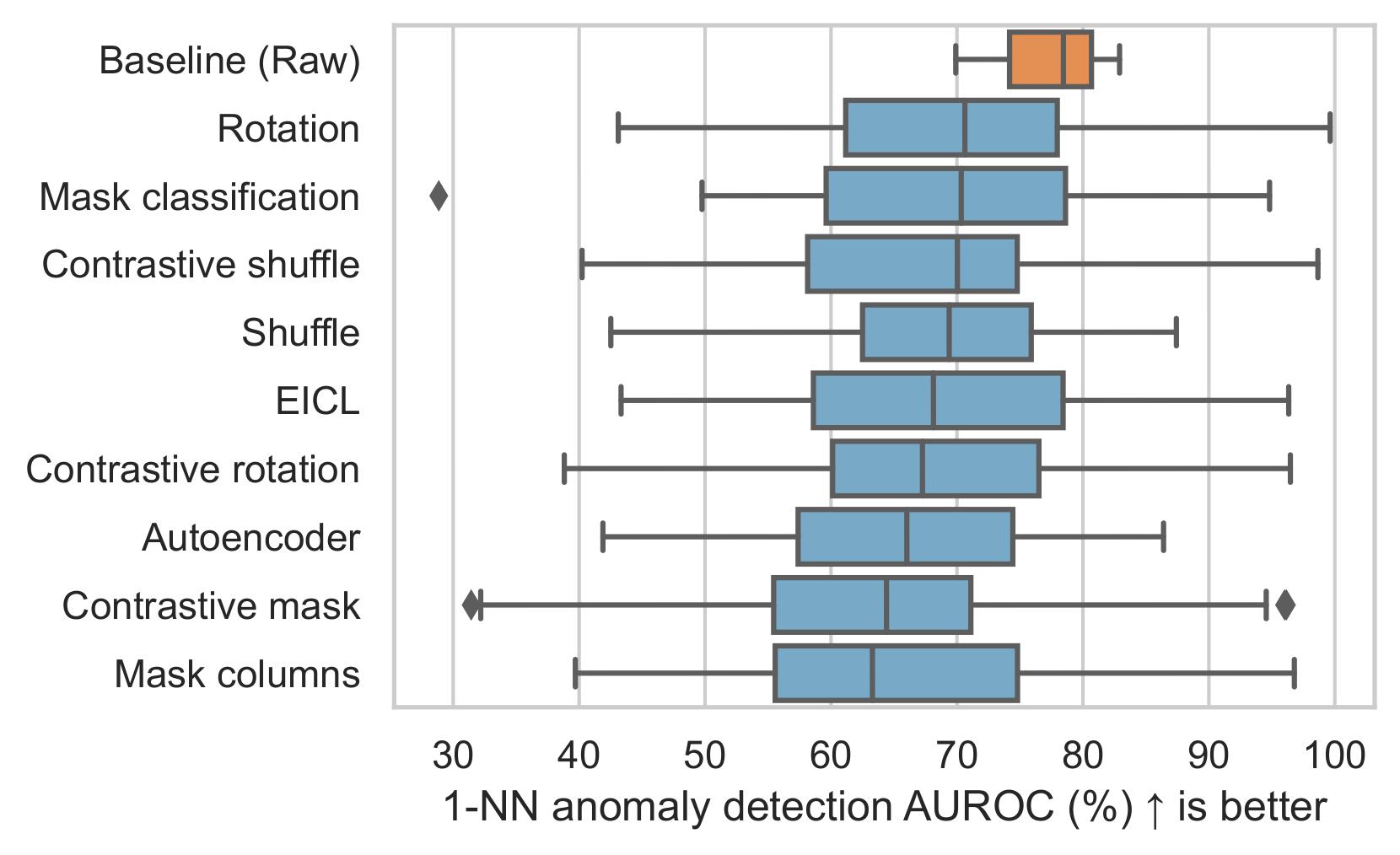}
         \caption{90\% removed entries}
     \end{subfigure}
     \hfill
     \begin{subfigure}[b]{0.45\linewidth}
         \centering
         \includegraphics[width=\textwidth]{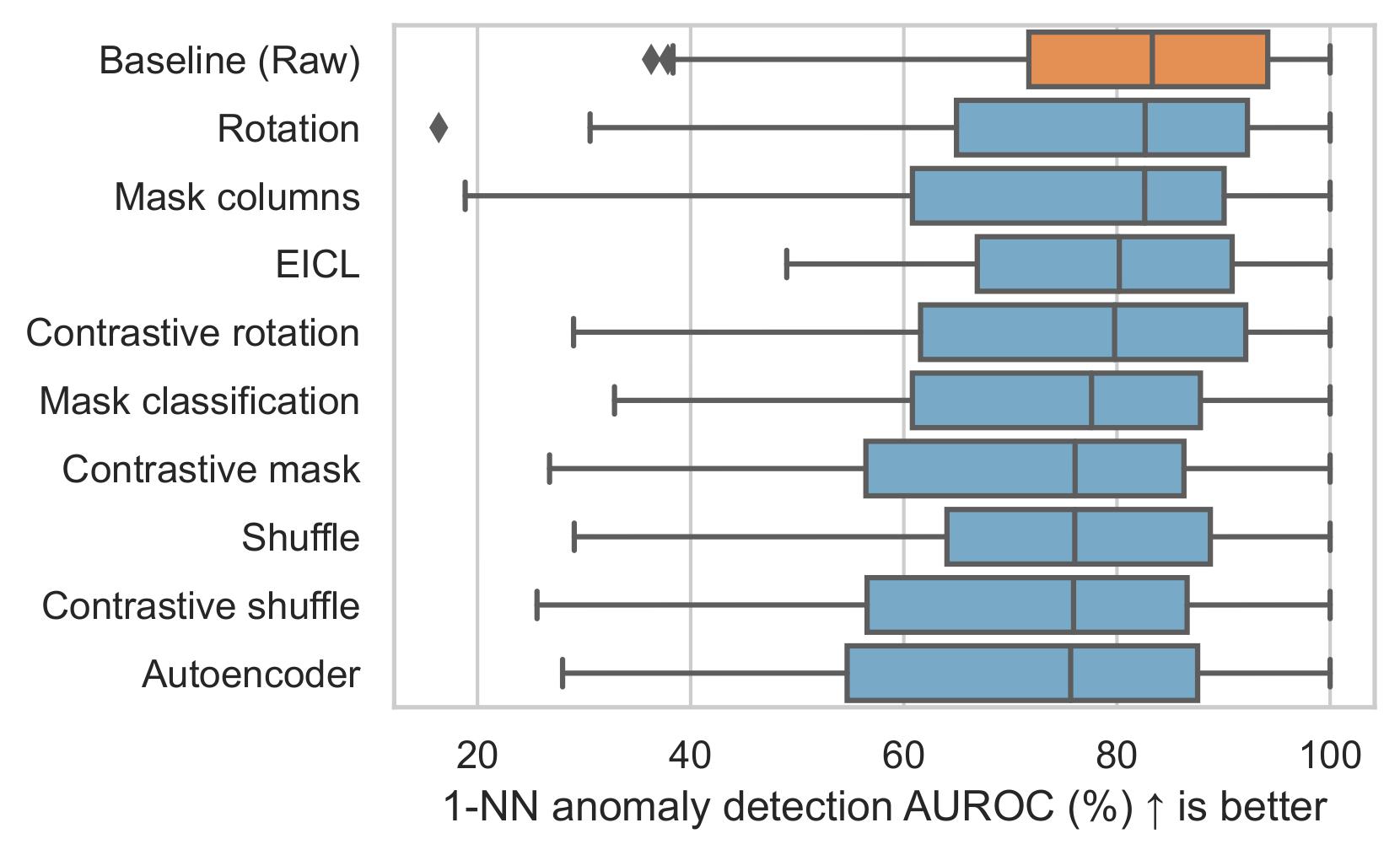}
         \caption{90\% removed features}
     \end{subfigure}
     \hfill
    \begin{subfigure}[b]{0.45\linewidth}
         \centering
         \includegraphics[width=\textwidth]{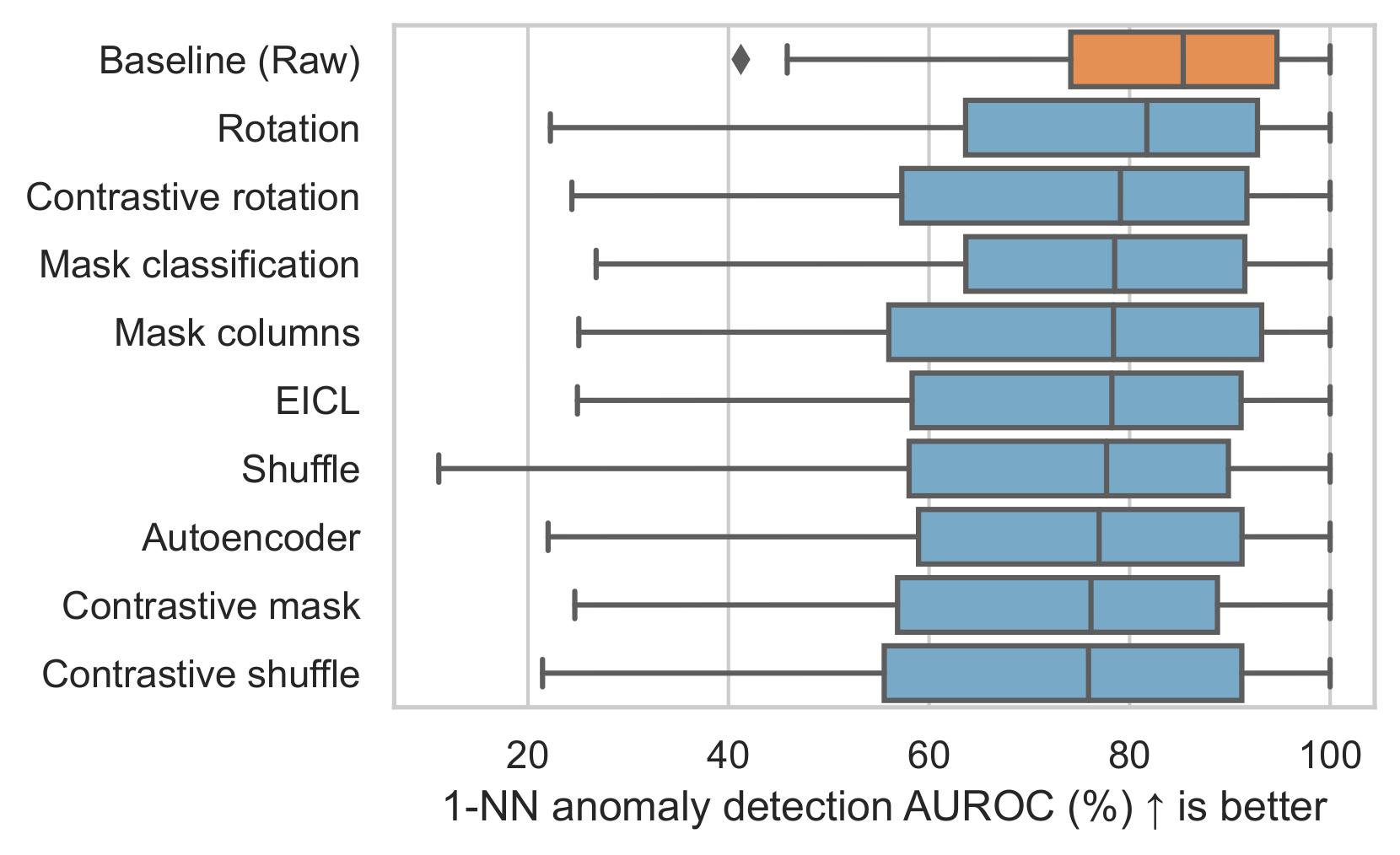}
         \caption{10\% selected features}
     \end{subfigure}
        \caption{Box plot comparing nearest neighbour AUROCs for each of the self-supervised pretext tasks on corrupted input data.}
        \label{fig:corrupt_ssl}
\end{figure}

\begin{figure}[h!]
     \centering
     \begin{subfigure}[b]{0.3\textwidth}
         \centering
         \includegraphics[width=\textwidth]{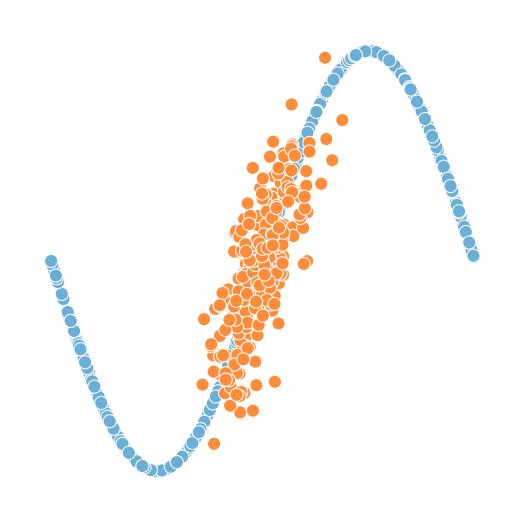}
         \caption{Curve}
     \end{subfigure}
     \hfill
     \begin{subfigure}[b]{0.3\textwidth}
         \centering
         \includegraphics[width=\textwidth]{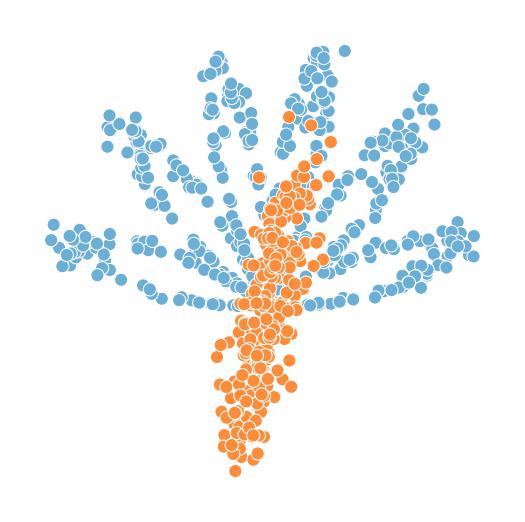}
         \caption{Flower}
     \end{subfigure}
     \hfill
     \begin{subfigure}[b]{0.3\textwidth}
         \centering
         \includegraphics[width=\textwidth]{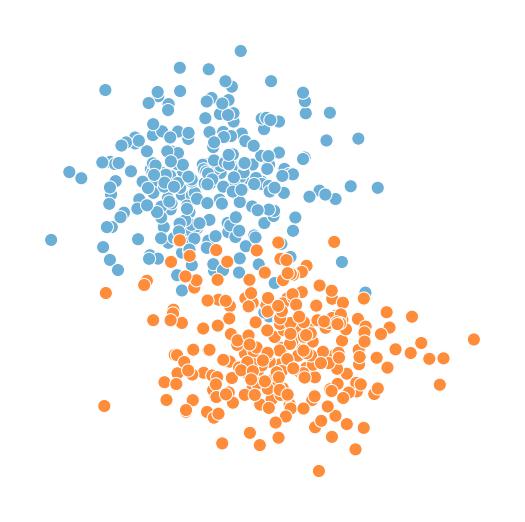}
         \caption{Gaussians}
     \end{subfigure}
     \hfill
    \begin{subfigure}[b]{0.3\textwidth}
         \centering
         \includegraphics[width=\textwidth]{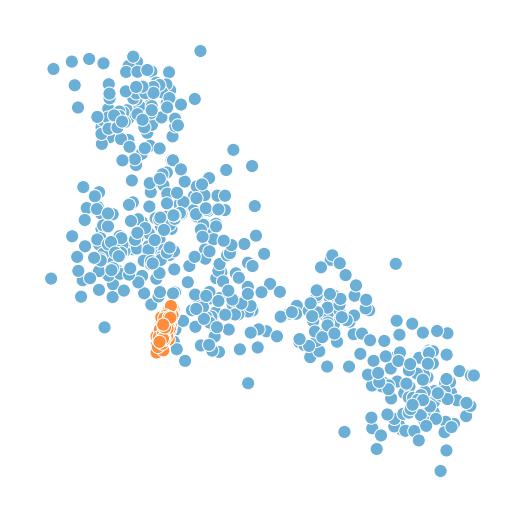}
         \caption{Multiple Gaussians}
     \end{subfigure}
          \hfill
    \begin{subfigure}[b]{0.3\textwidth}
         \centering
         \includegraphics[width=\textwidth]{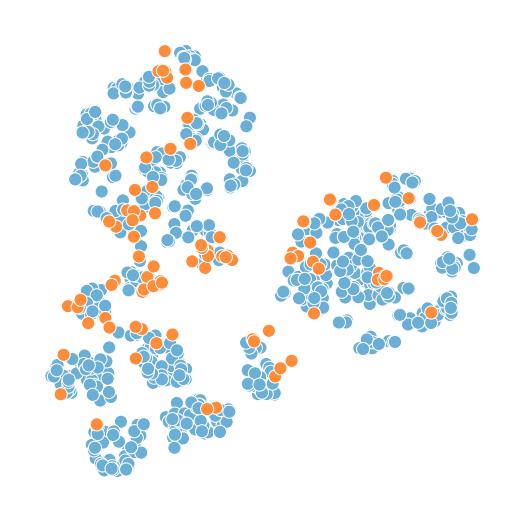}
         \caption{T-SNE projection of \textit{Letter} from ODDS}
     \end{subfigure}
          \hfill
    \begin{subfigure}[b]{0.3\textwidth}
         \centering
         \includegraphics[width=\textwidth]{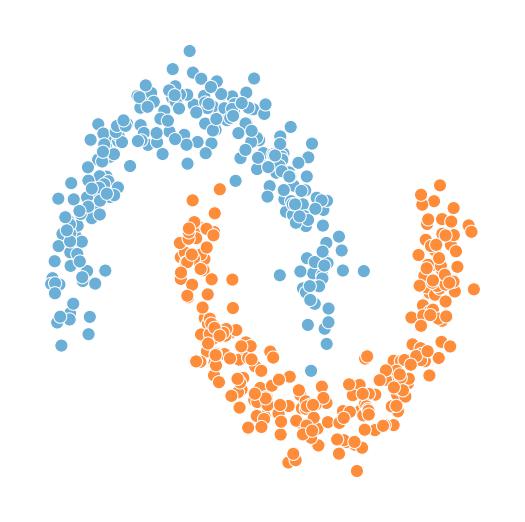}
         \caption{Moons}
     \end{subfigure}
          \hfill
    \begin{subfigure}[b]{0.3\textwidth}
         \centering
         \includegraphics[width=\textwidth]{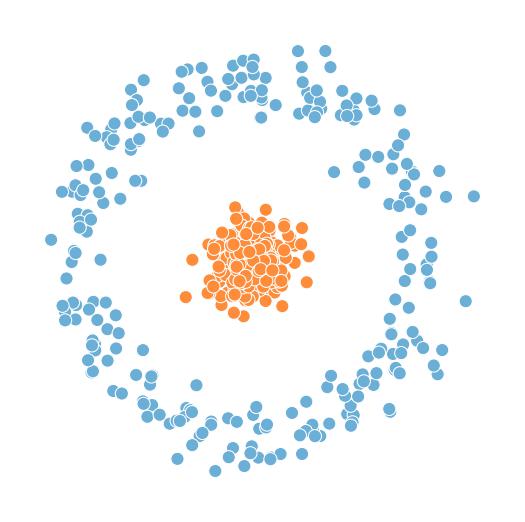}
         \caption{Ring}
     \end{subfigure}
          \hfill
    \begin{subfigure}[b]{0.3\textwidth}
         \centering
         \includegraphics[width=\textwidth]{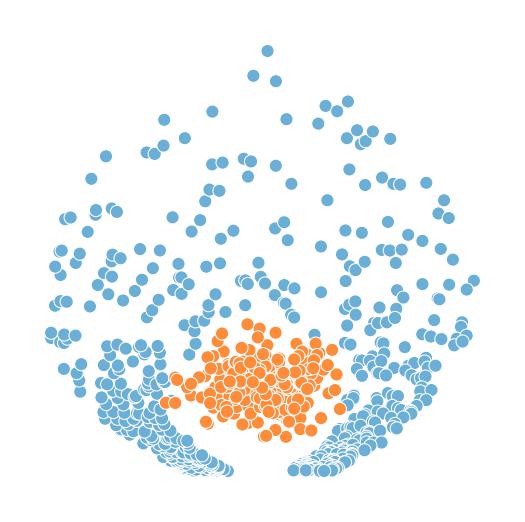}
         \caption{Pinched ring}
     \end{subfigure}
          \hfill
    \begin{subfigure}[b]{0.3\textwidth}
         \centering
         \includegraphics[width=\textwidth]{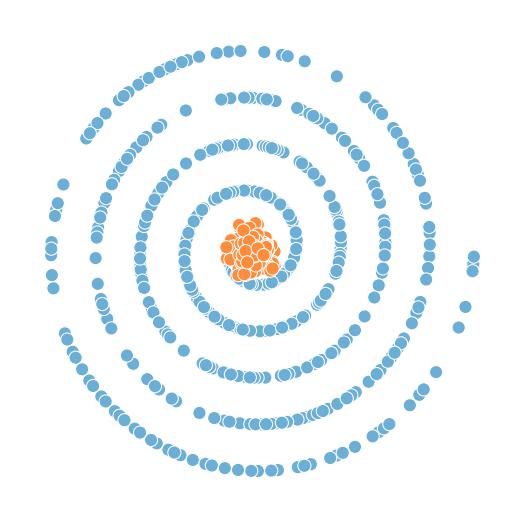}
         \caption{Spiral}
     \end{subfigure}
        \caption{Illustrations of the toy test data. Blue points are normal whereas orange points are anomalous.}
        \label{fig:toy_vis}
\end{figure}
\end{appendices}
\clearpage

\bibliography{sn-bibliography}

\end{document}